\documentclass{article} % For LaTeX2e
\usepackage{iclr2024_conference,times}

\usepackage{hyperref}
\usepackage{url}
\usepackage{csquotes}
\usepackage{placeins}

\usepackage{amsfonts}
\usepackage{amsmath}
\usepackage{amssymb}
\usepackage{amsthm}

\usepackage{cleveref} % for cref command
\usepackage{graphicx}
\usepackage{tcolorbox}

\usepackage{tikz}
\usepackage{booktabs}    % For better table lines
\usepackage{array}       % For improved table formatting
\usepackage{geometry}    % To adjust page margins if needed
\usepackage{caption}     % For better caption control
\usepackage{multirow}    % If multi-row cells are needed in future
\usepackage{adjustbox}   % To scale the table if needed
\usepackage{tabularx}
\usepackage{lscape}
\usetikzlibrary{trees,arrows.meta}

\usepackage{microtype}
\usepackage{graphicx}
\usepackage{subfigure}
\usepackage{booktabs} % for professional tables

\usepackage{hyperref}

\usepackage{hyperref}
\usepackage{url}
\usepackage{amsmath,amssymb}
\usepackage{amsthm}
\usepackage{leftidx}
\usepackage{graphicx}
\usepackage{fancyhdr}
\usepackage{enumitem}
\usepackage{tcolorbox}

\usepackage{algorithm}
\usepackage{algorithmicx}
\usepackage{algpseudocode}

\crefalias{section}{appendix}

\usepackage[textsize=tiny]{todonotes}

\title{IGDA: Interactive Graph Discovery through Large Language Model Agents}

\author{Alex Havrilla$^{1,2,*}$, David Alvarez-Melis$^{1,3}$, Nicolo Fusi$^{1}$}

\newcommand{\commentout}[1]{}

\iclrfinalcopy % Uncomment for camera-ready version, but NOT for submission.
\begin{document}

% Fabrizio comments: make a grid/taxonomy of synthetic data generation methods
% how does cost factor in as a function of the number of inference steps

% Other comments: dive deeper on some aspects of synthetic data generation. 
% how to encourage more exploration/diversity by combining with exterior

\maketitle

\vspace{-0.7cm}

\hspace{0.1cm} Microsoft Research$^{1}$, Georgia Tech$^{2}$, Harvard University$^{3}$

\renewcommand{\thefootnote}{}
\footnotetext{$^*$Work done during internship}

\vspace{0.5cm}

\begin{abstract}
  Large language models (\textbf{LLMs}) have emerged as a powerful method for discovery. Instead of utilizing numerical data, LLMs utilize associated variable \textit{semantic metadata} to predict variable relationships. Simultaneously, LLMs demonstrate impressive abilities to act as black-box optimizers when given an objective $f$ and sequence of trials. We study LLMs at the intersection of these two capabilities by applying LLMs to the task of \textit{interactive graph discovery}: given a ground truth graph $G^*$ capturing variable relationships and a budget of $I$ edge experiments over $R$ rounds, minimize the distance between the predicted graph $\hat{G}_R$ and $G^*$ at the end of the $R$-th round. To solve this task we propose \textbf{IGDA}, a LLM-based pipeline incorporating two key components: 1) an LLM uncertainty-driven method for edge experiment selection 2) a local graph update strategy utilizing binary feedback from experiments to improve predictions for unselected neighboring edges. Experiments on eight different real-world graphs show our approach often outperforms all baselines including a state-of-the-art numerical method for interactive graph discovery. Further, we conduct a rigorous series of ablations dissecting the impact of each pipeline component. Finally, to assess the impact of memorization, we apply our interactive graph discovery strategy to a complex, new (as of July 2024) causal graph on protein transcription factors, finding strong performance in a setting where memorization is impossible. Overall, our results show IGDA to be a powerful method for graph discovery complementary to existing numerically driven approaches.
\end{abstract}

\section{Introduction}
\label{sec:intro}

Given a set of variables $X_1,...,X_n$, the \textit{graph discovery} task involves finding a graph $G^*$ on the nodes $X_1,...,X_n$ whose  edges capture causal relationships between the \textit{parent} (source) and \textit{child} (target). Often, observational data can be collected for the variables $X_1,...,X_n$. This data can then be used to predict an initial graph $G_0$ using statistical causal discovery techniques \citep{Spirtes2016}. Recently, large language models (\textbf{LLMs}) have emerged as a competitive alternative method for predicting causal graphs \citep{kıcıman2024causalreasoninglargelanguage, abdulaal2024causal,chen2024clearlanguagemodelsreally}. Unlike pre-existing statistical methods, LLMs require no observational data \citep{kıcıman2024causalreasoninglargelanguage}, instead relying purely on semantic metadata such as variable names and descriptions. Another related line a work \citep{yang2024largelanguagemodelsoptimizers} investigates the abilities of LLMs to act as in-context black-box optimizers. Given an objective function $f$ and an evaluation budget $B$, the LLM  is tasked with finding a maximizer $x^*$ of $f$ by sequentially proposing queries $\{x_i\}_{i=1}^B$ and observing their associated values $\{f(x_i)\}_{i=1}^B$. Taken together, these directions suggest a powerful new application of LLMs: \textit{interactive graph discovery}.

Given an initial predicted graph $\hat{G}_0$ and a series of experiment rounds $1,...,R$, the interactive graph discovery problem involves minimizing some distance $d(\hat{G}_k, G^*)$ between the predicted graph $\hat{G}_k$ at round $k$ and the true   graph $G^*$ (unknown to the learner) through a  sequence of targeted experiments on edges $e = (X, Y)$ testing the effect of the parent variable $X$ on the child variable $Y$. The edge experiment operation is kept purposefully abstract, requiring only that binary feedback be given indicating the presence or absence of an edge. In practice this operation can be implemented via any number of experimental procedures (e.g. via hard interventions in the formal causal sense \citep{pearl_causal} or empirical methods such as randomized controlled trials \citep{sibbald1998understanding}). The IGD problem setup captures the process researchers go through everyday when designing and prioritizing experiments, guided by their prior experience, to study numerous potential relationships between any number of variables.

%\dam{Somehwere here or in Method, we need to define what we mean by `intervention`. Causal inference has various notions of intervention: hard (set/do), soft (probabilistic), stochastic, instrumental variable, etc. But in the actual experiments, my understanding is that rather than `intervening` the edges, their existence/non-existence is \textit{revealed}. How do we reconcile these two different paradigms?}. 

The interactive graph discovery problem requires the agent to solve two key sub-tasks:
\begin{enumerate}[topsep=0pt, parsep=0pt, itemsep=2pt, partopsep=0pt, left=2pt]%,leftmargin=*]
    \item \textbf{Experiment selection}: Selecting which edges $(X_i, X_j)$ to target for experimentation in the next round.
    \item \textbf{Graph updates}: Updating the predicted   graph from $\hat{G}_{k-1}$ to $\hat{G}_k$ given binary feedback based on the outcome of the previous experiments.
\end{enumerate} 

We propose to solve this task with the Interactive Graph Discovery Agent (\textbf{IGDA}): a novel LLM agent uncertainty-driven approach as an alternative to existing statistical methods \citep{olko2024trustnablagradientbasedintervention,scherrer2022learningneuralcausalmodels}. While statistical models can work well in some settings, they crucially rely on the abundance of domain specific observational and interventional numerical data. For many problems, such data might be hard or impossible acquire. LLMs, however, potentially contain relevant latent knowledge derived from vast amounts of variable semantic metadata contained in their pre-training or internet corpora. Further, we find that, via a combination of broad background knowledge and reasoning abilities, advanced LLMs \citep{grattafiori2024llama3herdmodels} are capable of updating their predictions and confidences when presented with experimental feedback revealing unexpected relationships between a subset of edges. This makes LLM based approaches a powerful alternative to statistical methods when numerical data is not available. 

In particular, IGDA predicts and maintains uncertainty estimates for each unknown edge $e \in \hat{G}_k$. Edges are then selected for experimentation by prioritizing those with the highest uncertainty. When feedback is received on the selected edges, pairwise-local updates on both edge predictions and uncertainty estimates are performed for each edge in $\hat{G}$ sharing a parent or child variable with an experimented edge. This process continues for $R$ rounds with $I$ edges selected for experimentation each round. We benchmark IGDA on eight real world   graphs, finding  uncertainty driven selection with local updates outperforms baselines. In summary, we make the following contributions:

\begin{itemize}[topsep=0pt, parsep=0pt, itemsep=2pt, partopsep=0pt, left=2pt]
    \item The interactive graph discovery problem as a novel setting for evaluating LLM capabilities.
    \item LLM-based uncertainty-guided edge experiment selection as a policy for prioritizing edge experimentation.
    \item A local update strategy for robustly updating the predicted graph $G_k$ with binary experiment feedback.
    \item Ablations rigorously evaluating the contribution of each pipeline component and other discovery strategies.
\end{itemize}

\section{Background and Related Work}

\paragraph{Causal Discovery and LLMs.} The causal discovery task involves learning causal relationships from observed empirical data \citep{peters, Spirtes2016}. Many proposed algorithms exist \citep{Spirtes1993CausationPA, yu2019daggnndagstructurelearning,Nauta2019CausalDW,zheng2018dagstearscontinuousoptimization,Chickering2002OptimalSI} attempting to solve the causal discovery problem. However, these methods are known to struggle on real world graphs where observations are noisy or common structural assumptions are violated \citep{chevalley2023causalbenchlargescalebenchmarknetwork,tu2019neuropathicpaindiagnosissimulator}. 

Recently, LLMs have emerged as an alternative approach to causal discovery \citep{kıcıman2024causalreasoninglargelanguage,abdulaal2024causal,vashishtha2023causalinferenceusingllmguided,li2024realtcdtemporalcausaldiscovery,lampinen2023passivelearningactivecausal}. \citet{kıcıman2024causalreasoninglargelanguage} first investigated the capability of LLMs to act as zero-shot causal discovery agents using only semantic information and pairwise prompting on each variable pair. Follow-up work \citep{abdulaal2024causal} further improves LLM predictions with observational data by selecting for predictions which maximize data likelihood. \citet{vashishtha2023causalinferenceusingllmguided} utilize \textit{triplet prompting} to prevent cycles when the causal graph is acyclic. They show only a topological ordering on variables is required for many common causal reasoning tasks \citep{chu2023causaleffectestimationrecent}. Other works \citep{zhou2024causalbenchcomprehensivebenchmarkcausal, chen2024clearlanguagemodelsreally} benchmark LLMs across a range of causality related tasks including causal discovery and causal inference confirming that LLMs struggle with integrating numerical data. 

Another line of work more related to our proposed interactive causal discovery problem  studies how to incorporate background knowledge into causal discovery algorithms \citep{meek2013causalinferencecausalexplanation}. Define a set of \textit{background knowledge} as the tuple $\mathcal{K} = (F,R)$, where $F$ specifies a set of ``forbidden'' graph edges and $R$ specifies a set of ``required'' graph edges. \citet{meek2013causalinferencecausalexplanation} presents an algorithm for constructing a causal graph consistent with $\mathcal{K}$ by leveraging an assumed structural directed acyclic graph (DAG) property. Building on \citet{meek2013causalinferencecausalexplanation}, \citet{Chickering2002OptimalSI} proposes a greedy search algorithm that performs well in practice.

Most related are statistical methods from the causal discovery literature which aim to efficiently choose a sequence of interventions to discover causal structure \citep{scherrer2022learningneuralcausalmodels,olko2024trustnablagradientbasedintervention}. In particular, Gradient based Interventional Targeting (\textbf{GIT}) \citep{olko2024trustnablagradientbasedintervention} utilizes existing neural causal discovery methods \citep{lippe2022efficientneuralcausaldiscovery} to learn a distribution over possible graph structures and variable assignments. For each round of intervention, GIT prioritizes variables whose simulated interventional distribution have large gradient with respect to the structural training loss.

In contrast to these works, our proposed algorithm utilizes LLMs to reason about the semantic/physical, as opposed to formal/structural, relationships between variables and edges in causal graphs. For this reason we are not required to make any structural assumptions on an underlying DAG, as is common in the causal discovery literature. This is desirable as in practice many real-world causal graphs are cyclic and poorly structured \citep{brain_cg, arctic}. Additionally our method does not rely on observational or interventional data for real world graphs which may be expensive to acquire but crucial for good performance with statistical methods.

\paragraph{LLMs as Optimizers.} Another growing line of work utilizes LLMs as black-box optimizers \citep{yang2024largelanguagemodelsoptimizers, roohani2024biodiscoveryagentaiagentdesigning}. \citet{yang2024largelanguagemodelsoptimizers} introduce the notion of an LLM as a generic optimizer and use it to optimize performance objectives stemming from a range of tasks including linear regression and mathematical word problems \citep{cobbe2021trainingverifierssolvemath}. Other works \citep{madaan2023selfrefineiterativerefinementselffeedback, havrilla2024glorewhenwhereimprove} examine the self-refinement capabilities of LLMs where the LLM must reason and self-improve on earlier responses. A growing number of papers apply LLMs to optimal experiment design and discovery \citep{roohani2024biodiscoveryagentaiagentdesigning, ai4science2023impactlargelanguagemodels,gao2024empoweringbiomedicaldiscoveryai,majumder2024discoverybenchdatadrivendiscoverylarge,jansen2024discoveryworldvirtualenvironmentdeveloping}. \citet{roohani2024biodiscoveryagentaiagentdesigning} apply LLMs to gene discovery tasks which aim to find highly-influential parent genes affecting the regulation of a downstream target gene. \citet{majumder2024discoverybenchdatadrivendiscoverylarge,jansen2024discoveryworldvirtualenvironmentdeveloping} both present benchmarks evaluating the ability of LLMs to perform real-world and synthesized discovery tasks.

%Misc: 
%\citep{lyle2023discobaxdiscoveryoptimalintervention}

\section{Method}
\label{sec:method}

\begin{figure*}
    %\hspace{0.15cm}
    \includegraphics[scale=0.18]{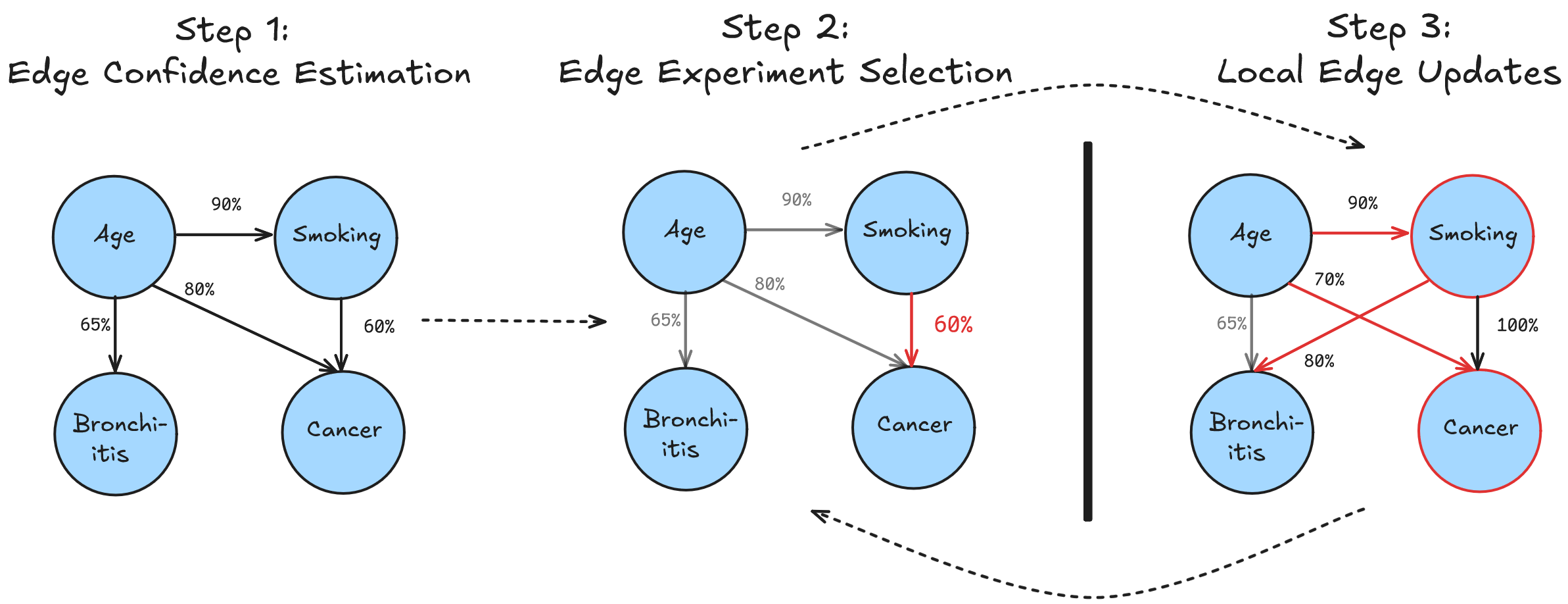}
    \caption{Diagram of the interactive graph discovery process through LLMs. The process begins by predicting edges and confidences for each edge. Interactive discovery then proceeds by selecting the most uncertain edges for experimentation. The LLM then updates its predictions and confidences for edges adjacent to the selected edge. Note: only edges predicted as present are shown.}
    \label{fig:icdllm}
\end{figure*}

%\dam{This might be an overkill for a workshop paper, but I believe the method section would be simpler/more concise if we adopt the causal inference terminology and definitions. A side product of this is that if anyone from that community ends up reviewing the paper, they'll appreciate the familiarity. For example, we can define $Y\rightarrow X$ if $Y$ is a causal parent of $X$. Then, the set of casual parents of a variable $X$ as $\textup{Pa}(X) = \{X_i \mid X_i \rightarrow X\}$. With this, we can define the ground truth graph as having edges $(X_i,X_i)$ iff $X_i\in \textup{Pa}(X_j)$. }

%\alex{See update.}

\paragraph{Setup.} As input we are given a set of variables $X_1,...,X_n$  with associated metadata including variable names and variable descriptions. We use the notation $Y \to X$ to indicate when variable $Y$ has a direct effect on variable $X$ and the set of parents of a variable $X$ as $Pa(X) = \{X_i : X_i \to X\}$. We can then consider the directed ground truth graph $G^* = \{(X_i, X_j) : X_i \in Pa(X_j)\}$ with unlabeled and unweighted edges. The only assumed graph structure is simplicity i.e. no self-edges or multi-edges. No additional structure on the graph (such as acyclicity) is assumed. We can frame the prediction of $G^*$ as an edge-wise binary classification problem over the complete graph $K_n$, where an edge $(X_i, X_j)$ has the label $l_{ij} = 1$ if $X_i \to X_j$ and $l_{ij} = 0$ otherwise. $G^*$ can then be written as a collection of ground truth labelings $G^* = \{(X_i, X_j, l_{ij}) : 1 \leq i \neq j \leq n\}$. 

The \textit{interactive graph discovery} task then aims to learn $G^*$ by interacting with the discovery environment via \textit{experiments} on each edge $(X_i, X_j)$. We define an \textit{experiment} on an edge $(X_i, X_j)$ as an operation revealing the ground truth label $l_{i,j}$. This experiment operation is purposefully kept abstract for generality and could correspond to any number of real-world experimental experimental strategies including formal do operations \citep{pearl_causal} or empirical randomized control trials \citep{sibbald1998understanding}. Interactive graph discovery then proceeds in two phases:

\begin{itemize}[label={}]
    \item \textbf{Phase 1 (Zero-shot prediction):} Produce an initial   graph prediction $\hat{G}_0$ using available variables $X_1,...,X_n$ plus semantic metadata.
    \item \textbf{Phase 2 (Interactive Discovery):} Over a series of $R$ rounds, propose $I$ edge experiments on $(X_i, X_j)$ each round and receive binary feedback on $l_{ij}$. Use this to produce an updated prediction $\hat{G}_{r-1} \to \hat{G}_r$
\end{itemize}

We evaluate the accuracy of a prediction $\hat{G}$ using the F1 objective, i.e.
\begin{equation*}
    F1(G^*, \hat{G}) = \frac{2\cdot \texttt{Precision}_{\hat{G}}\cdot \texttt{Recall}_{\hat{G}}}{\texttt{Precision}_{\hat{G}} + \texttt{Recall}_{\hat{G}}}
    \vspace{-0.2cm}
\end{equation*} where $\texttt{Precision}_{\hat{G}}$ and $\texttt{Recall}_{\hat{G}}$ are computed with the label predictions $(X_i, X_j, \hat{l}_{ij}) \in \hat{G}$ and $l_{ij}$ as ground truth. The goal of the interactive discovery process is then to maximize $F1(G^*, \hat{G}_R)$.

\begin{algorithm}[ht]
\caption{Interactive Graph Discovery Through LLMs}
\label{alg:main_llm_discovery}
\begin{algorithmic}
\Procedure{LLMGraphDiscovery}{$\hat{G}$, $R$, $I$} \Comment{$\hat{G}$: initial graph (confidences), $R$: rounds, $I$: experiments/round}
    \For{$r \gets 1$ to $R$}
        \Comment{Step 1: select edges for experimentation}
        \State $sorted\_edges \gets \text{sort}(\hat{G}, \text{key}=\text{``conf''})$
        \State $experiments \gets sorted\_edges[1:I]$ % Python slicing notation is better expressed this way
        \State $binary\_feedback \gets \text{do\_experiments}(experiments)$
        \Comment{Step 2: update $\hat{G}$ using feedback}
        \For{$i \gets 1$ to $I$}
            \State $edge, edge\_gt \gets experiments[i], binary\_feedback[i]$
            \State $adjacent\_edges \gets \text{get\_adjacent\_edges}(edge)$
            \For{$a \gets 1$ to $\text{length}(adjacent\_edges)$}
                \State $\hat{G}[a][\text{``update\_confs''}] \gets \text{LLMLocalUpdate}(edge, edge\_gt, a)$
            \EndFor
        \EndFor
        \For{$a$ in $\hat{G}$} % or \ForAll{$a$ in $\hat{G}$} if you want to emphasize all
            \State $\hat{G}[a][\text{``conf''}] \gets \text{mean}(\hat{G}[a][\text{``update\_confs''}])$
            \State $\hat{G}[a][\text{``pred''}] \gets \mathbf{1}_{\hat{G}[a][\text{``conf''}] > 0}$ % Use \mathbf for bold 1
        \EndFor
    \EndFor
    \State \textbf{return} $\hat{G}$ % Or \Return $\hat{G}$ for more algorithm-specific return
\EndProcedure
\end{algorithmic}
\end{algorithm}

\paragraph{Method.} Our proposed method IGDA begins by generating a zero-shot graph prediction $\hat{G}_0$. A prediction for each variable pair $(X_i, X_j)$, $1 \leq i \neq j \leq n$, is generated by prompting an LLM to reason about $X_i \to X_j$ in a manner similar to the pairwise-prompting strategy utilized in \citet{kıcıman2024causalreasoninglargelanguage}. In addition, we prompt the LLM to reason about its confidence in the prediction and output a confidence score from 1 - 100. Section \ref{sec:prompts} shows the exact prompt used. To obtain a reliable confidence estimate we sample the LLM $K = 16$ times. We denote the initial confidence for $(X_i, X_j)$ as $c_{ij}^0$ and set it to be the (signed) average over $K = 16$ output confidences. The initial edge label $l_{ij}^0$ is then taken as the boolean $l_{ij}^0 = \textbf{1}_{c_{ij}^0 \geq 0}$. This gives us the initial prediction $\hat{G}_0$. 

Next, in each experimentation round $r \leq R$, we sort the confidence scores $\{c_{ij}^r : 1 \leq i,j\leq n\}$ by absolute value and experiment on the $I$ edges with the lowest absolute confidence (and highest uncertainty). This reveals the ground truth labels $l_{ij}$ for each experimented edge $(X_i, X_j)$. Using this feedback, we update the predicted edge labels for experimented edges to $l_{ij}^{r+1} = l_{ij}$ and the confidences to $c_{ij}^{r+1} = 100$. Additionally, we prompt the LLM, conditioned on the ground truth label $l_{ij}$, to update its prediction and confidence for each edge $(X_i, X_k)$ or $(X_l, X_j)$, $1 \leq k,l \leq n$ which shares a node with $(X_i, X_j)$ and has absolute confidence less than 100. We call each update to an edge $(X_l, X_k)$ a \textit{local update}. It may be that an edge $(X_l, X_k)$ is adjacent to multiple experimented edges $(X_{i_1},X_{j_1}), (X_{i_2},X_{j_2})$ in a single round and thus receives multiple local updates. To manage these cases we set the next confidence $c_{lk}^{r+1}$ to the (signed) average of all individual local updates to $c_{lk}^r$. Then we set $l_{lk}^{r+1} = \textbf{1}_{c_{lk} \geq 0}$ as before. This continues until the final round $R$ is reached. 

We call the complete discovery pipeline the \textit{Interactive Graph Discovery Agent} (IGDA). A diagram of the full pipeline is shown in Figure \ref{fig:icdllm}. We report all prompts in \cref{sec:prompts}.

\section{Results}
\label{sec:results}

We evaluate our approach on seven real-world graphs. The graphs range in size from 8 to 30 nodes (variables) and vary widely in structure (some are acyclic while others are cyclic). Details for each graph can be found in Appendix~\ref{sec:graphs}. To produce initial zero-shot graph predictions $\hat{G}_0$ for all graphs we utilize pairwise causal prompting as in \citet{kıcıman2024causalreasoninglargelanguage} with \texttt{Meta-Llama-3-70B-Instruct} \citep{grattafiori2024llama3herdmodels} as the base LLM. We chose \texttt{Meta-Llama-3-70B-Instruct} as at the time of our experiments it was the best open-source model with advanced reasoning and instruction following capabilities For the interactive discovery phase we then initialize all methods using $\hat{G}_0$. We compare our method against several baselines:

\begin{itemize}[label={}]
    \item \textbf{Random selection:} Starting from $\hat{G}_0$ we randomly select edges for experimentation. After receiving binary feedback we update incorrect predictions on experiment edges for the next round. We do not allow edges to be selected for experimentation twice.
    \item \textbf{Static confidence selection:} We select edges for experimentation based on the initial confidence scores $c_{ij}$. No updates are performed beyond fixing incorrect predictions in the experimentation set.
    \item \textbf{Gradient-based Intervention Targeting (GIT):} We adapt the statistical GIT method \citep{olko2024trustnablagradientbasedintervention} by selecting the node at each round which has a) not already been selected and b) has the largest loss gradient under a neural causal model \citep{lippe2022efficientneuralcausaldiscovery} trained with all available observational and interventional training data. We initially train the model with 5000 observational datapoints sampled from the ground-truth graph. 100 additional interventional datapoints on the experiment node are sampled from the ground-truth graph and added to the training set after each round of experimentation.
\end{itemize}

\begin{figure*}
    \centering
    \includegraphics[width=0.25\linewidth]{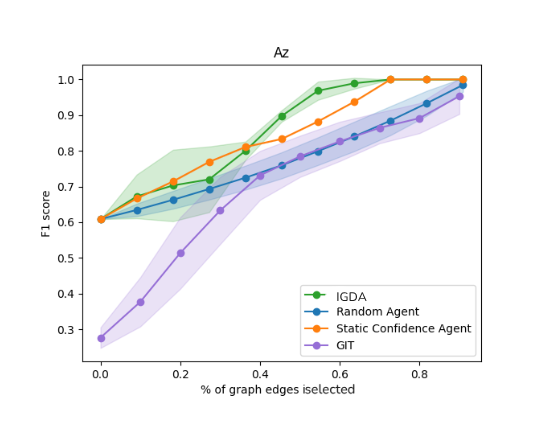}
    \includegraphics[width=0.22\linewidth]{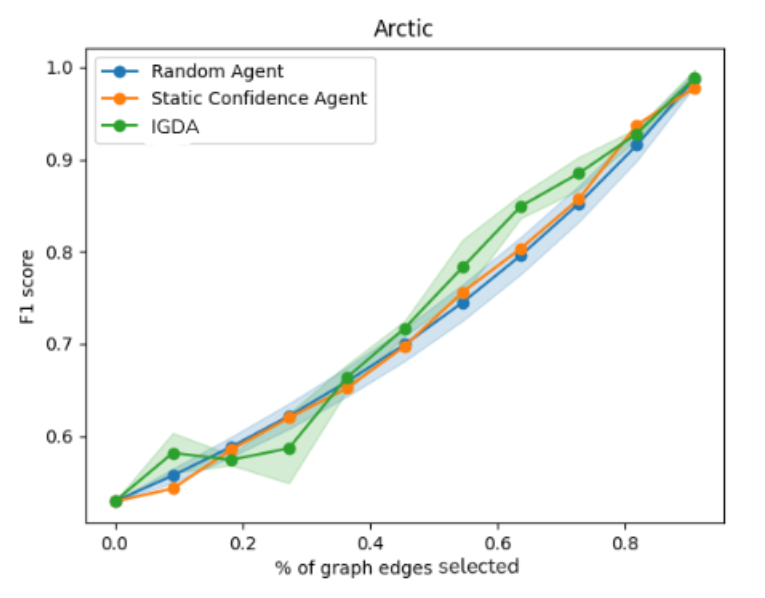}
    \includegraphics[width=0.25\linewidth]{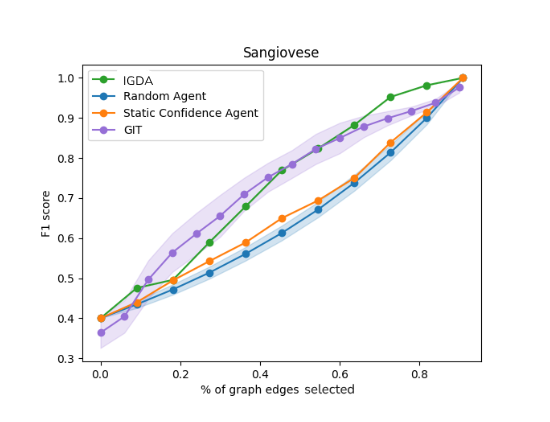}

    \includegraphics[width=0.24\linewidth]{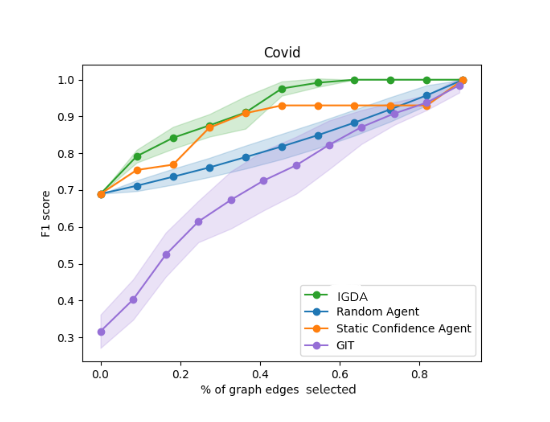}
    \includegraphics[width=0.24\linewidth]{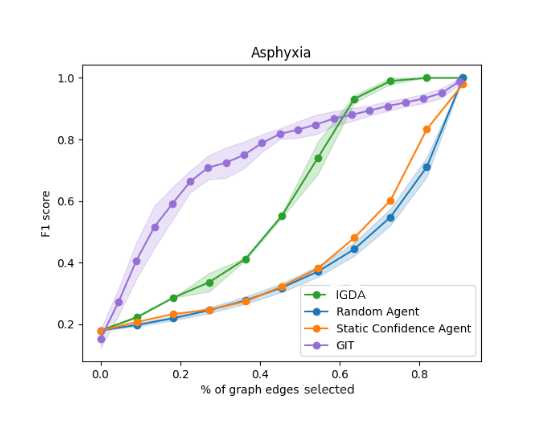}
    \includegraphics[width=0.24\linewidth]{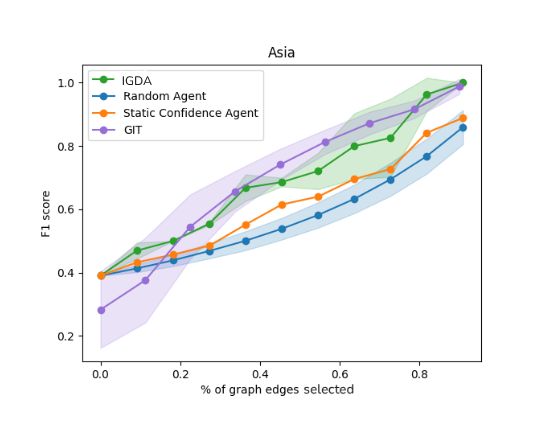}
    \includegraphics[width=0.24\linewidth]{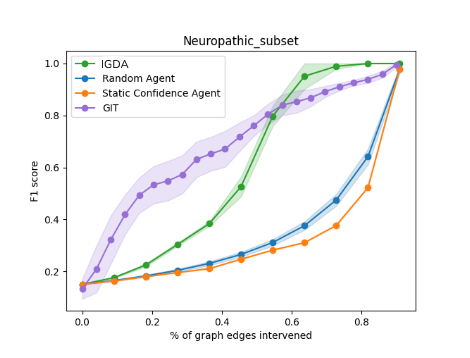}
    
    \caption{Results on real world graphs showing F1 score of the predicted graph against percentage of edges in the graph selected. IGDA almost always outperforms both the random baseline and static selection via uncertainty. Note: static confidence selection without local updates is deterministic and thus has no confidence intervals. Additionally, GIT is not reported on the Arctic graph because the grpah is cyclic.}
    \label{fig:results}
\end{figure*}

\texttt{
Meta-Llama-3-70B-Instruct}  is used as the base LLM when applicable. To assess performance, we plot the mean F1 score, averaged over five independent runs, against the percentage of edges selected in each graph. Results are shown in Figure \ref{fig:results}.

\begin{figure}
    \centering\includegraphics[width=0.5\linewidth]{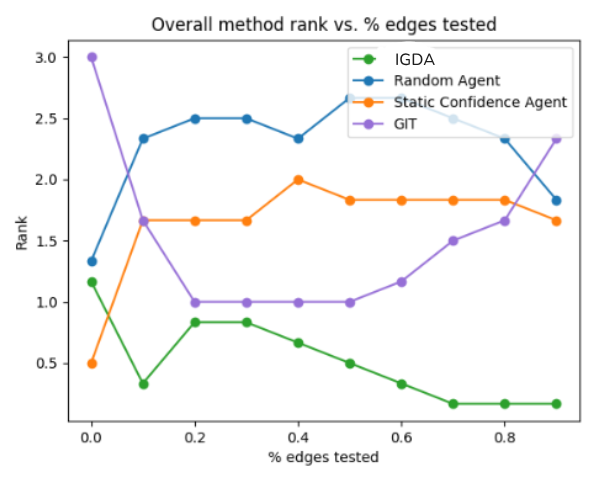}
    \caption{Average rank of each method when numbered from $0$ to $2$ across each timestep on each graph. The full LLM driven update agent consistently achieves rank $0$ across all timesteps. Note: \textbf{lower is better}.}
    \label{fig:ranks}
\end{figure}

\paragraph{Uncertainty driven experiment selection with local updates performs best.} Uncertainty driven experiment selection with the LLM utilizing experimental feedback for local updates performs best on nearly all graphs. Further, it outperforms the random selection baselines at nearly every round on every graph, at times by up to 0.5 absolute F1 score. The only exception to this is the Arctic sea ice graph where local updates initially perform poorly. We attribute this to the highly cyclic and thus harder-to-predict graph structure. Additionally, the method significantly outperforms the statistical GIT baseline on both Az and Covid graphs and remains competitive on the rest. Figure \ref{fig:ranks} plots the average rank of all methods over all timesteps, confirming IGDA's strong performance. Notably, even on graphs where the LLM proposes a poor zero-shot initial prediction, the LLM is able to recover quickly, converging to the correct   structure with local updates. This suggests the LLM is able to effectively utilize experiment feedback even when lacking detailed domain knowledge.

\paragraph{Local updates can outperform random selection even with few experiments.} Allowing the LLM to make local edge updates using experiment feedback quickly improves the predicted graph even when relatively few edges are selected. This behavior is particularly desirable, as in practice it may be expensive to experiment on even a small fraction of all edges. On some graphs, where the initial LLM confidence estimates are good, the static confidence selection baseline without local updates is also able to quickly outperform random selection. Yet, even when the initial confidence estimates are subpar, local updates compensate and allow for the prediction to quickly improve with just a few edge experiments. This again demonstrates the broad effectiveness of local updates even when initial predictions are poor.

\paragraph{Static uncertainty driven selection performs better than random selection.} Despite not fully utilizing experimental feedback, static uncertainty driven selection still outperforms the random selection baseline on five out of seven graphs. This method performs particularly well on AZ and Covid graphs where the initial LLM predictions are already reasonably good. On these graphs static uncertainty selection quickly outperforms randomly selection and is competitive even with local updates. This shows that, on a subset of the graphs, the LLM's confidence in its predictions are well-calibrated, allowing our selection policy to prevent wasting experiments on edges which are most likely already correct. However, we also see the LLM's confidence estimates can be poorly calibrated on graphs for which the initial predictions are inaccurate. See for example the Asphyxia and Neuropathic pain graphs, which start with initial F1 score less than 0.2. On these graphs the static confidence selection component struggles to outperform the random baseline. 

\begin{figure}
    \centering
    \includegraphics[width=0.5\linewidth]{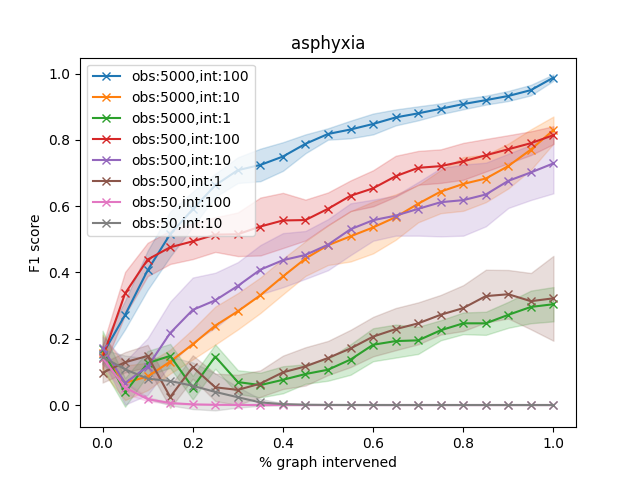}
    \caption{GIT with varying amounts of observational and interventional data. Decreasing either observational or interventional sample sizes can decrease performance by over 0.2 F1 score.}
    \label{fig:git_asphyxia}
\end{figure}

\paragraph{GIT performance heavily depends on availability of both observational and interventional data} With ample data (5000 observational samples and 100 interventional samples per node) the statistical GIT methods performs well on most graphs where it is applicable (i.e. the graph is acyclic). However, we find this good performance heavily depends on the availability of such data, with decreases in both observational and interventional sample sizes significantly impacting results. In Figure \ref{fig:git_asphyxia} we plot the performance of GIT on the Asphyxia graph with varying amounts of data demonstrating this effect. Results on more graphs are presented in the appendix. In contrast, IGDA does not depend at all on the availability of numerical observational or interventional data. Instead, IGDA relies on the complementary availability of semantic metadata of graph variables within either its pretraining dataset or on the internet.

In an effort to better understand the factors behind IGDA's success we conduct a number of ablations in the following section.

\subsection{Ablations}
\label{sec:ablations}

\begin{figure*}
    \centering
    \includegraphics[width=0.24\linewidth]{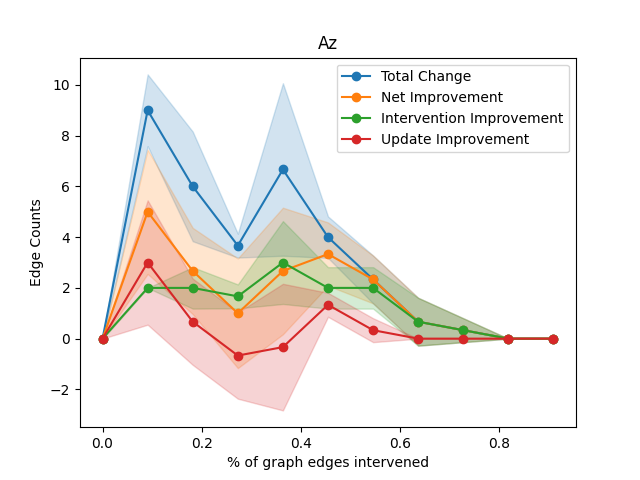}
    \includegraphics[width=0.24\linewidth]{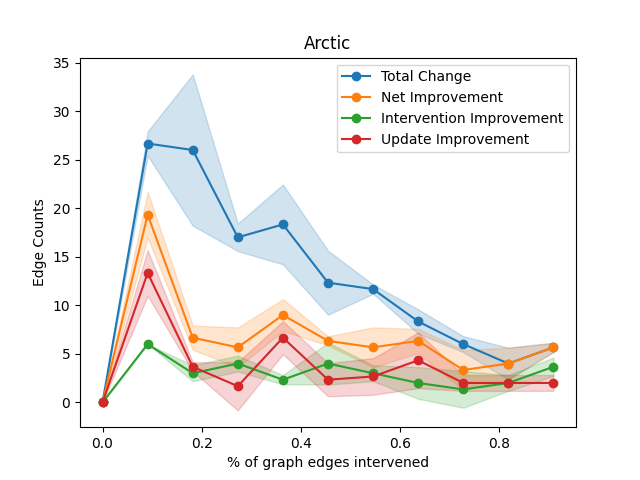}
    \includegraphics[width=0.24\linewidth]{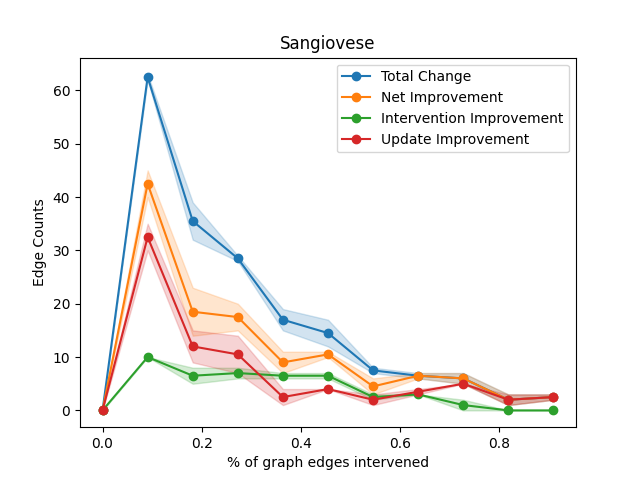}
    \includegraphics[width=0.24\linewidth]{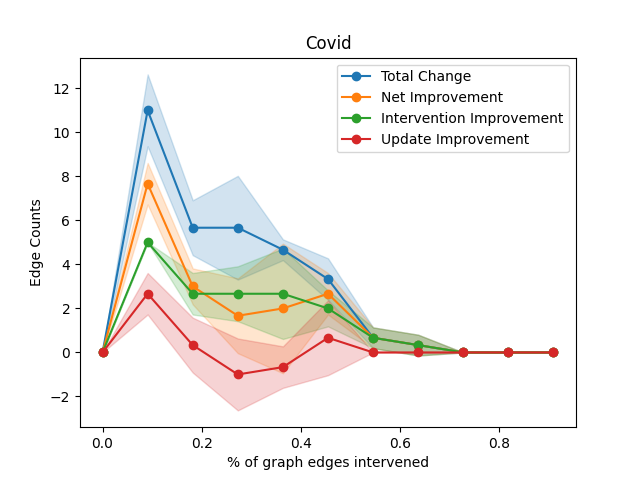}

    \includegraphics[width=0.24\linewidth]{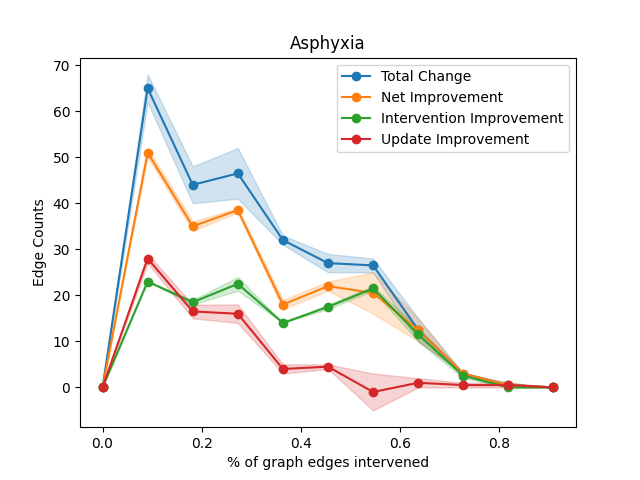}
    \includegraphics[width=0.24\linewidth]{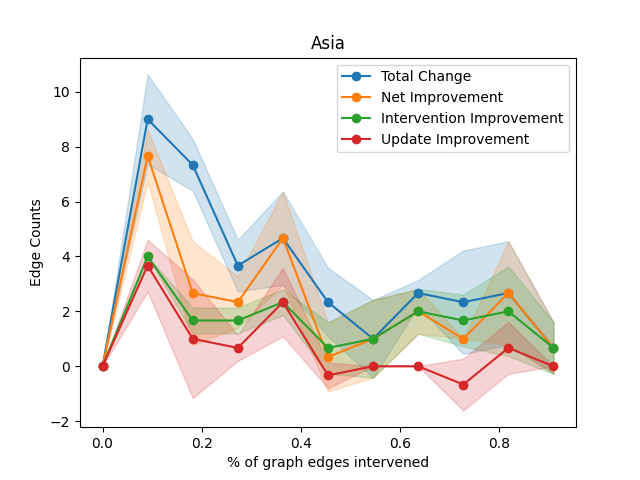}
    \includegraphics[width=0.24\linewidth]{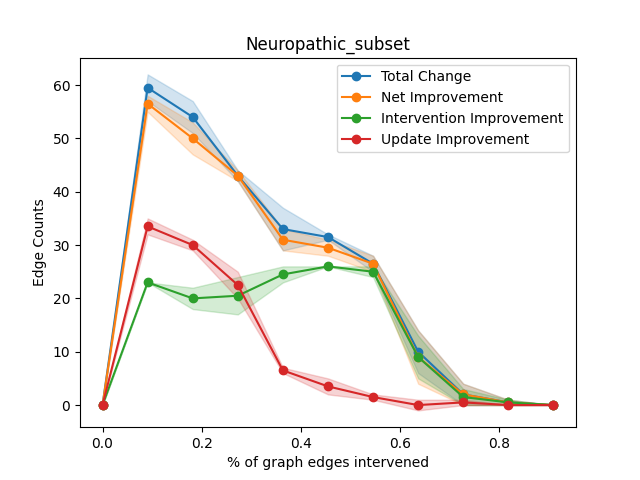}
    \includegraphics[width=0.24\linewidth]{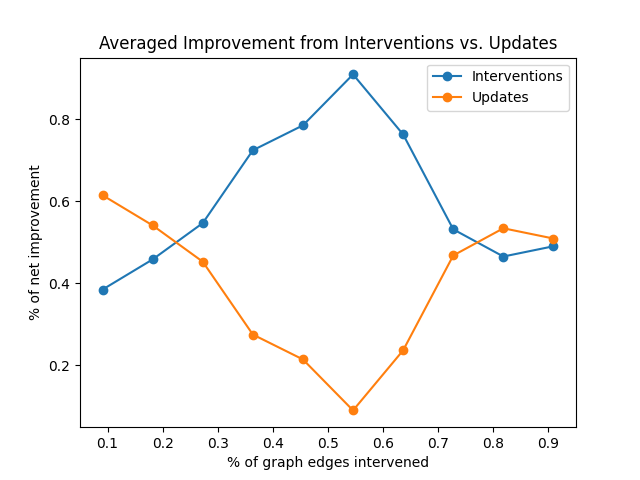}
    \caption{\% Improvement from experiments vs. LLM prediction updates across timesteps. Improvement directly from LLM updates peaks early but then falls off. Improvement from experiments stays constant or improves with more experiments as confidence scores become better calibrated.}
    \label{fig:improvement_analysis}
\end{figure*}

\paragraph{Impact of experiment improvements versus update improvements} As a starting point we define the $\textit{net graph improvement}$ in a round $r$ as the difference between the number of edges correctly classified in in $\hat{G}_r$ versus in $\hat{G}_{r-1}$. If an edge $(X_i, X_j)$ is correctly classified in $\hat{G}_r$ but not in $\hat{G}_{r-1}$ we say it has been $\textit{improved}$. Recall there are two potential mechanisms of improvement for $(X_i, X_j)$: 1) $(X_i, X_j)$ was selected for experimentation in the previous round $r-1$ and feedback on the experiment was received at the start of round $r$ 2) sThe prediction for $(X_i, X_j)$ was updated by the LLM after receiving experiment feedback for an adjacent edge $(X_k, X_l)$. We call the former improvements $\textit{experiment improvements}$ and the latter $\textit{update improvements}$. In a given round $r$ we are interested in how much of the net improvement for a graph is due to experiment improvements versus update improvements. To examine this, we plot both quantities in Figure \ref{fig:improvement_analysis} for the discovery processes discussed in the previous section. In addition, we plot the net graph improvement and total number of edges changed from each round.

In all seven graphs we see both the total number of changed edges and the net improved edges peak at the first round and then decay towards zero. Notably, on some graphs there is a significant gap between net improvement and total change, indicating many edges changed during dynamic updates are misclassified after previously being correctly classified. This decline in total and net change is reflected in the number of update improvements which peak early and sharply decline to zero. This observation supports our intuition above that allowing the LLM to dynamically update edge predictions without direct experimental feedback on the edge can dramatically improve performance at small percentages of experiments. In contrast, experiment improvement accounts for a smaller percentage (less than 40\%) of edge improvements early on. However, in most graphs the number of experiment improvements stays nearly constant until at least 50\% of edges are already selected. As a result, improvement from experiments grows to account for 90\% of all edge improvements for rounds performed during this period. This demonstrates improvements from experiment and updates complement each other, with \textbf{update improvement driving net improvement early and experiment improvement driving net improvement later on}. 

Our analysis here also confirms the effectiveness of allowing the LLM agent to update both the prediction \textbf{and} confidence for an edge. Even when only considering improvements from experiments when doing local updates, we see a major improvement over the static confidence baseline. This suggests the \textbf{updates made to edge confidence scores are equally important in achieving good performance}, allowing for sustained experiment improvement throughout the discovery process.

\begin{figure}[t]
    \centering
    \includegraphics[width=0.32\linewidth]{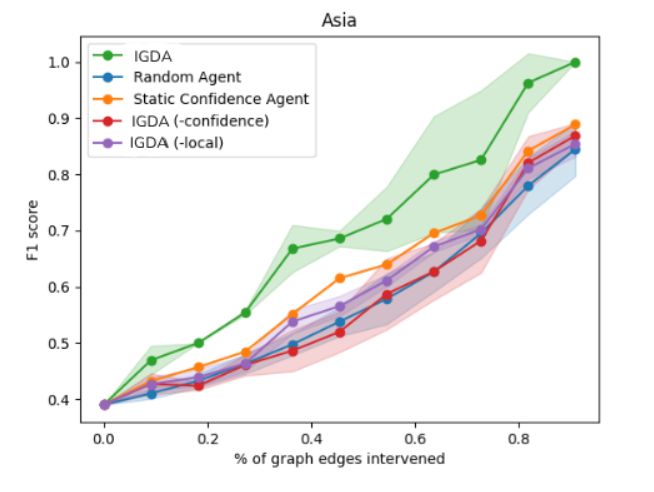}
    \includegraphics[width=0.32\linewidth]{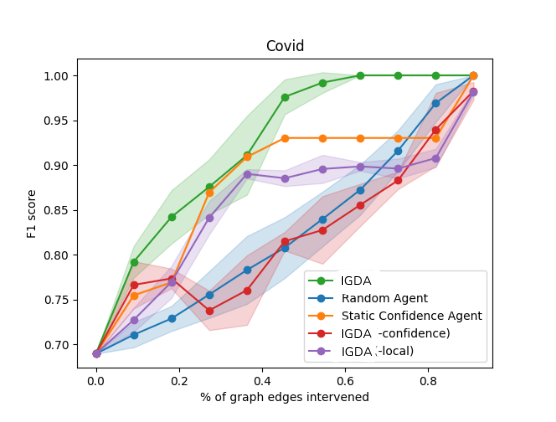}
    \includegraphics[width=0.32\linewidth]{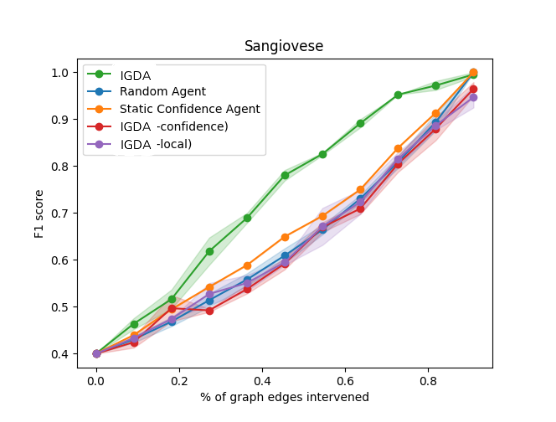}
    \caption{Ablating confidence based edge selection and local update prompting.}
    \label{fig:prompt_ablations}
\end{figure}

\paragraph{Impact of Confidence Based Selection and Local Prompting} We now ablate the impact of two key components of our discovery strategy: 1) confidence based edge selection and 2) local update prompting. To ablate 1) we directly prompt the LLM to generate a list of edges to experiment on instead of selecting via confidence. This requires us to put the entire current predicted graph $\hat{G}_r$ in-context. When dynamically updating $\hat{G}_r$ after receiving experimental feedback we remove all confidence estimates but retain the local prompting strategy. To ablate 2) we retain the same confidence edge selection proposed but replace local update prompts after with a single global update prompt containing the current prediction $\hat{G}_r$ and all recently received experiment feedback. We report the results of running the interactive discovery process with these methods in Figure \ref{fig:prompt_ablations}.

%TODO: better motivate why IGD is an important problem. Talk about science and prioritizing experiments

We find both ablations struggle to perform better than the random baseline. Local updates without confidence selection perform well early on but fall off quickly. F1 score on the Covid graph even regresses after the initial improvements, likely due to incorrect local updates and a poor experiment selection policy. This suggests in addition to providing a strong experiment selection procedure, maintaining running confidence estimates for each edge reduces the variance of local updates from experiment feedback. Turning to the ablation for local prompting, we again find performance not much better than the random baseline. Surprisingly, even on Covid where the static confidence selection performs well, confidence based selection + global updates still struggles. This indicates the base LLM is not able to correctly update the predicted graph when giving everything in context at once. This further motivate the practical importance of the local prompting procedure, which greatly simplifies the context the LLM must consider in each model call. Additionally, we note that for large enough graphs, putting everything in context is simply not feasible. By contrast, local prompting is easily scalable to larger graphs, albeit at a quadratic cost.

\begin{figure}
    \centering
    \includegraphics[width=0.24\linewidth]{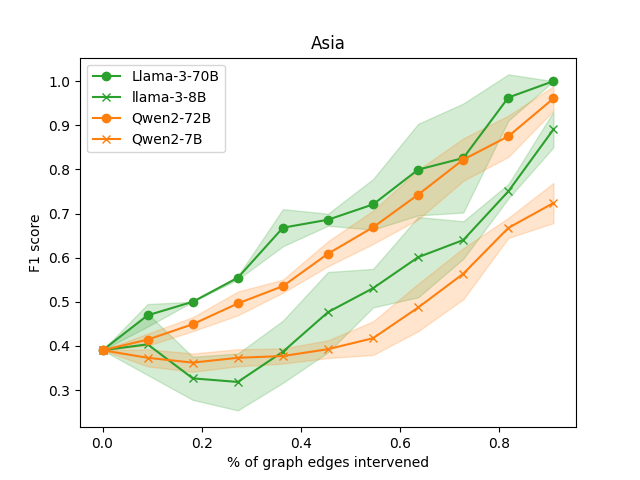}
    \includegraphics[width=0.24\linewidth]{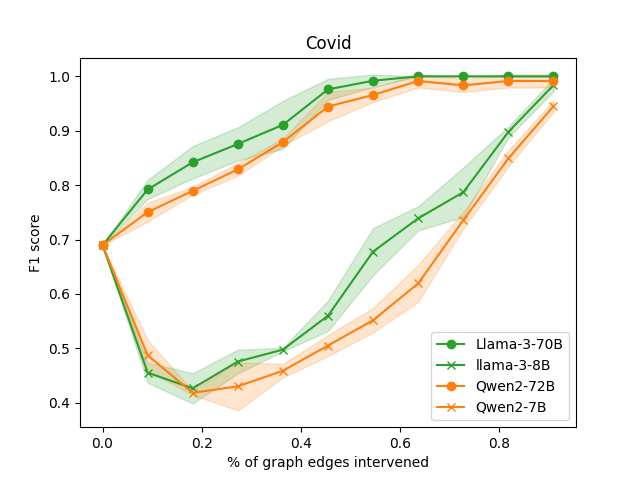}
    \includegraphics[width=0.24\linewidth]{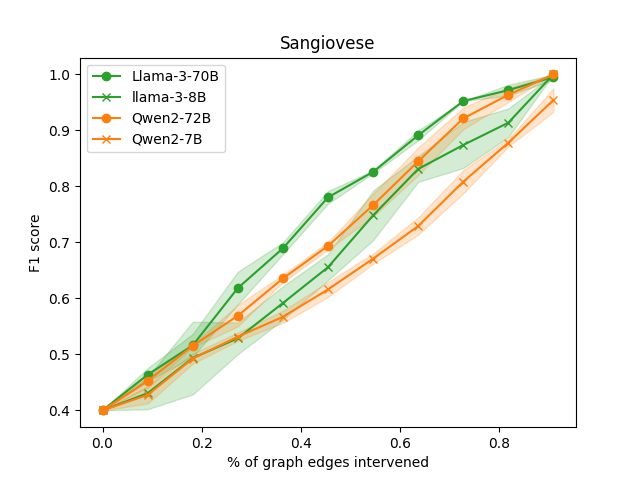}
    \includegraphics[width=0.24\linewidth]{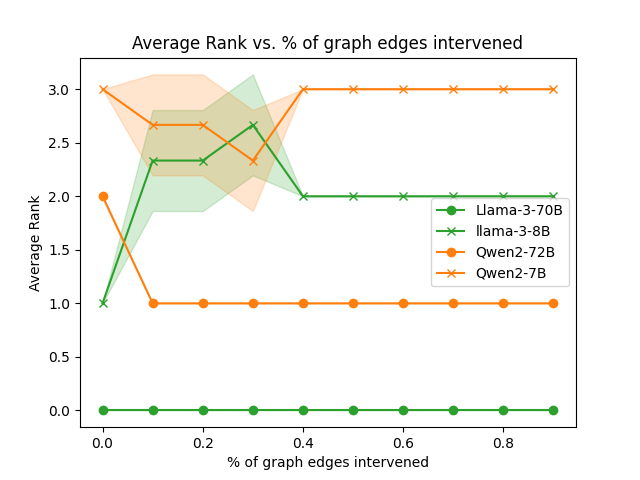}
    \caption{Performance of LLM driven interactive   discovery on different sized models. Small LLMs (8B params) underperform the random baseline. }
    \label{fig:model_ablations}
\end{figure}

\paragraph{Impact of the LLM Model Size} The above experiments exclusively use a single base LLM (\texttt{
Meta-Llama-3-70B-Instruct}) to perform both the initial round of zero-shot edge predictions and dynamically update edge predictions/confidences using experiment feedback. Now, we examine the impact of changing both the base model size and type. In Figure \ref{fig:model_ablations} we initialize the discovery process with zero-shot predictions made by \texttt{
Meta-Llama-3-70B-Instruct} and run local updates using the smaller \texttt{
Meta-Llama-3-8B-Instruct} as well as two models from the Qwen2 series. 

We find the original \texttt{
Meta-Llama-3-70B-Instruct} consistently performs best on all graphs at every time step. The other 70B model, \texttt{
Qwen2-72B-Instruct}, performs similarly but consistently worse. In contrast, on the Asia and Covid   graphs, both 8B models perform worse than even the random baseline. Surprisingly \texttt{
Meta-Llama-3-8B-Instruct} performs reasonably well on the Sangiovese graph, performing similarly even to the 9x larger Qwen2 70B model. Overall however these results indicate performance on the interactive graph discovery task can be substantially improved with model scale. 

We next investigate the performance of different models on the initial zero-shot edge prediction task. Using the pairwise confidence estimation prompt in Section \ref{sec:prompts} we prompt each of four models to produce a zero-shot prediction $\hat{G}_0$ with edge confidence values. Using the predicted confidence estimates we run greedy static confidence selection procedure as in \ref{sec:results}. Ranks for each selection procedure averaged over all graphs are plotted in Figure \ref{fig:static_ranks}. F1 scores in each graph are reported in Figure \ref{fig:f1_static_confidence} in the Appendix.

\begin{figure}
    \centering
    \includegraphics[width=0.5\linewidth]{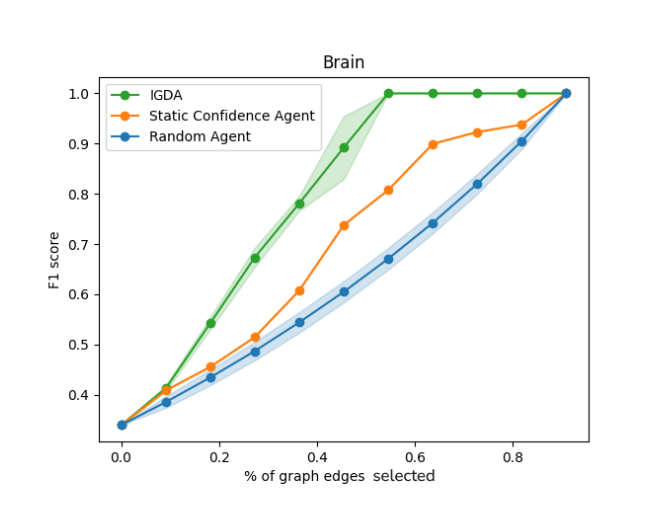}
    \caption{Performance curves of uncertainty driven selection + local prompting vs.~baselines on the Brain graph \citep{brain_cg} recently published in July 2024. Both LLM-driven methods perform well despite the graph not possibly being in the LLM's training data. Note: GIT is not reported because the graph is cyclic.}
    \label{fig:brain_f1}
\end{figure}

\paragraph{Impact of Memorization} The success of LLMs in   discovery stems from their immense background knowledge acquired during pre-training. This background knowledge informs the model during edge prediction and confidence calibration, allowing for strong performance even zero-shot. However, if benchmark graphs are contained verbatim in pre-training data, memorization becomes a significant confounding factor. To investigate to what extent memorization impacts performance we find a recently published   graph (published in July 2024) from \citet{brain_cg} modeling the gene regulatory network underlying 29 protein transcription factors. Because \texttt{Meta-Llama-3-70B-Instruct} finished training in 2023 this graph is guaranteed to be memorization free. Figure \ref{fig:brain_f1} plots the performance of uncertainty driven edge selection + local updates compared to the static selection and random baseline.

Figure \ref{fig:brain_f1} shows our confidence driven selection + local update approach performs very well even on graphs with minimal memorization contamination. As previously observed, local prediction updates allow for fast improvement over the random baseline even with a small number of experiments. Surprisingly, the static confidence selection approach also works well here. This indicates zero-shot edge confidence scores can be well calibrated on graphs with no contamination from memorization. We additionally note this graph has a complex   structure with many cycles of varying lengths. This shows our method performs well even on graphs which strongly violate often assumed DAG conditions.

\section{Conclusions and Future Work}
\label{sec:conclusion}

In this work we proposed IGDA as a novel  application of LLMs to interactive graph discovery. Our experiments confirm the proposed IGDA method significantly outperforms baselines. Our ablations confirm both uncertainty driven edge selection and local updates using experiment feedback as importantly contributing to the method's good performance. Further, this method is complementary to existing statisical methods for experiment design or causal discovery (e.g. GIT \citep{olko2024trustnablagradientbasedintervention}). Statistical methods utilize available observational/interventional numerical data to make predictions and confidence estimates whereas IGDA utilizes available variable semantic metadata to make predictions and confidence estimates. Designing a method leveraging both numerical and semantic variable data is promising future work.

\bibliography{refs}
\bibliographystyle{icml}

\newpage
\appendix

\section{Prompts}
\label{sec:prompts}

\begin{tcolorbox}[colback=yellow!30, colframe=red!40!black, title=Zero-shot Confidence Estimation Prompt]
\{task\_description\}
Your goal is to understand the direct causal parents of \{target\}.
Another variable is a direct causal parent of \{target\} if an experiment on the variable affects \{target\} \
and there are no other causal parents between the variable and \{target\}.
Now, you must determine whether \{parent\} is a causal parent of \{target\}.
Here is a list of all other variables to consider:

\{variables\_info\}

Do some brainstorming, comparing relevant characteristics of both variables \
and then print your judgment at the end of your response enclosed in the tags \
$<$decision$>$YES/NO$<$/decision$>$. Print YES if \{parent\} is causal. \
Otherwise print NO. \
You should also print your confidence from a scale from 1 - 100 (with 100 being most confident) \
in the tags $<$confidence$>$...$<$ /confidence$>$.

Information about \{target\}:
\{target\_info\}

Information about \{parent\}:
\{parent\_info\}
\end{tcolorbox}

\begin{tcolorbox}[colback=yellow!30, colframe=red!40!black, title=Parent Update Prompt]
You are a causal discovery expert. You have been \
given the following list of variables and tasked with predicting the \
true causal graph through a sequence of experiments on edges.

\{variables\_info\}

Note: each edge has an associated confidence value from 1 - 100. 
The presence of an edge is represented as (A$->$B,CONFIDENCE) where A is the parent and B is the child. 
The absence of an edge is represented as (NOT A$->$B, CONFIDENCE)

From one experiment you have discovered
\{experiment\_feedback\}
Previously you predicted
\{experiment\_prediction\}

Now you should update your belief about the other edges of \{parent\} \
based on the results of the experiment. Consider the predicted edge

\{other\_edge\_prediction\}

Now you should reason about how to update your belief about the above edge based on the experiment. \
This means you can either keep your confidence the same, update your confidence, or \
change your prediction entirely. \
At the end of your response give your updated prediction at the end of \
your response in the format
$<$decision$>$PARENT/NOT CAUSAL$<$/decision$>$ $<$confidence$>$CONFIDENCE$<$/confidence$>$.
Print 'PARENT' if the edge should be present and 'NOT CAUSAL' if the edge should be absent.

You should do this in three steps.

Step 1: Brainstorm what physical causal connection there may be, if any.

Step 2: Reason about what the experiment feedback tells you. \
Think carefully about how similar the new child is to the experimental child.

Step 3: Give your final decision.
\end{tcolorbox}

\begin{tcolorbox}[colback=yellow!30, colframe=red!40!black, title=Child Update Prompt]
You are a causal discovery expert. You have been \
given the following list of variables and tasked with predicting the \
true causal graph through a sequence of experiments on edges.

\{variables\_info\}

Note: each edge has an associated confidence value from 1 - 100. 
The presence of an edge is represented as (A$->B$,CONFIDENCE) where A is the parent and B is the child. 
The absence of an edge is represented as (NOT A$->$B, CONFIDENCE)

From one experiment you have discovered
\{experiment\_feedback\}
Previously you predicted
\{experiment\_prediction\}

Now you should update your belief about the other edges of \{child\} \
based on the results of the experiment. Consider the predicted edge

\{other\_edge\_prediction\}

Now you should reason about how to update your belief about the above edge based on the experiment. \
This means you can either keep your confidence the same, update your confidence, or \
change your prediction entirely. \
At the end of your response give your updated prediction at the end of \
your response in the format
$<$decision$>$PARENT/NOT CAUSAL$<$/decision$>$ $<$confidence$>$CONFIDENCE$<$/confidence$>$.
Print 'PARENT' if the edge should be present and 'NOT CAUSAL' if the edge should be absent.

You should do this in three steps.

Step 1: Brainstorm what physical causal connection there may be, if any.

Step 2: Reason about what the experiment feedback tells you. \
Think carefully about how similar the new parent is to the experiment parent.

Step 3: Give your final decision.
\end{tcolorbox}

\newpage

\section{GIT Ablations}

\begin{figure*}[ht]
    \centering
    \includegraphics[width=0.32\linewidth]{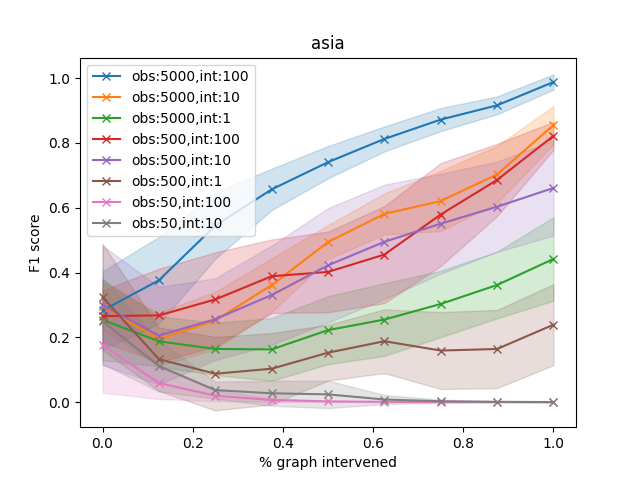}
    \includegraphics[width=0.32\linewidth]{figs/stat_ablations/asphyxia.png}
    \includegraphics[width=0.32\linewidth]{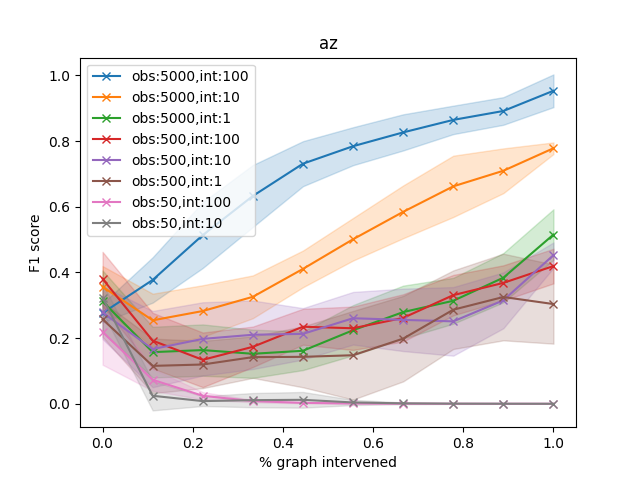}
    \\
    \includegraphics[width=0.32\linewidth]{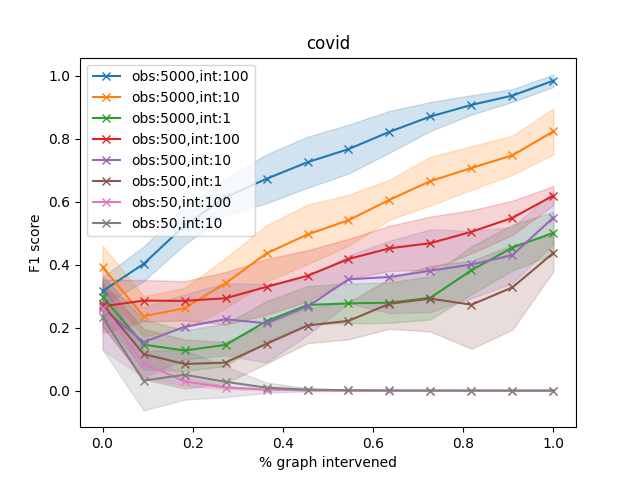}
    \includegraphics[width=0.32\linewidth]{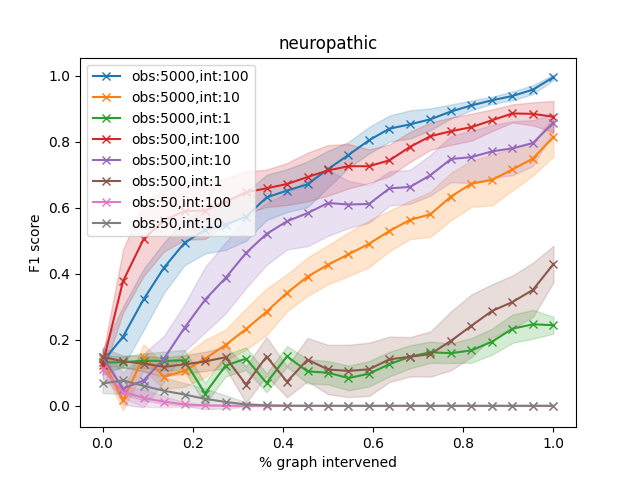}
    \includegraphics[width=0.32\linewidth]{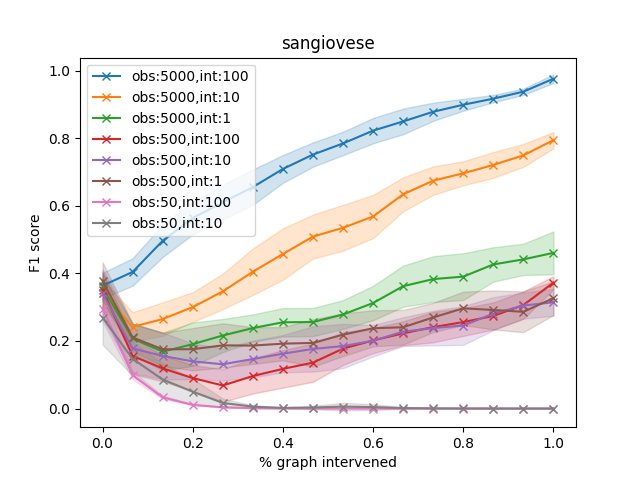}
    \caption{GIT ablations with varying amounts of observational and interventional data.}
    \label{fig:git_ablations}
\end{figure*}

Figure \ref{fig:git_ablations} plot GIT performance \citep{olko2024trustnablagradientbasedintervention} over six causal graphs with varying amounts of observational and interventional data.

\section{Static Confidence Selection over Multiple Models}

\begin{figure}[ht]
    \centering
    \includegraphics[width=0.32\linewidth]{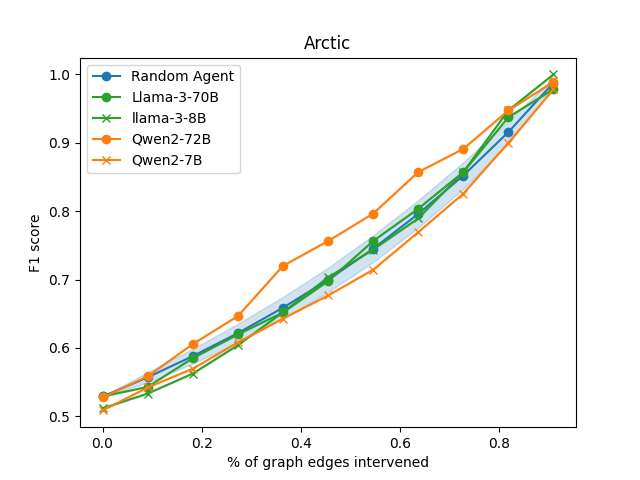}
    \includegraphics[width=0.32\linewidth]{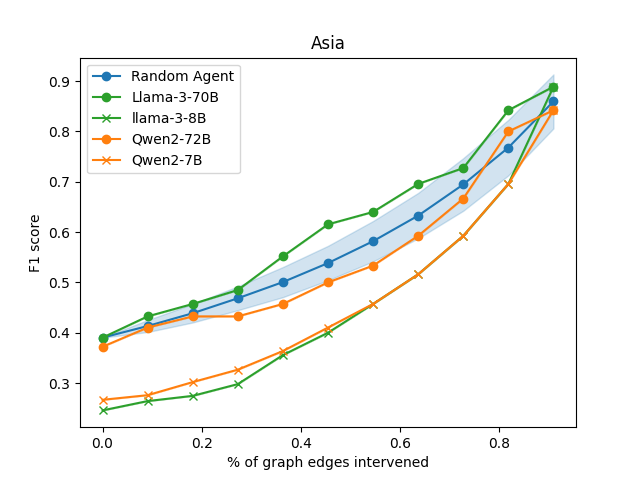}
    \includegraphics[width=0.32\linewidth]{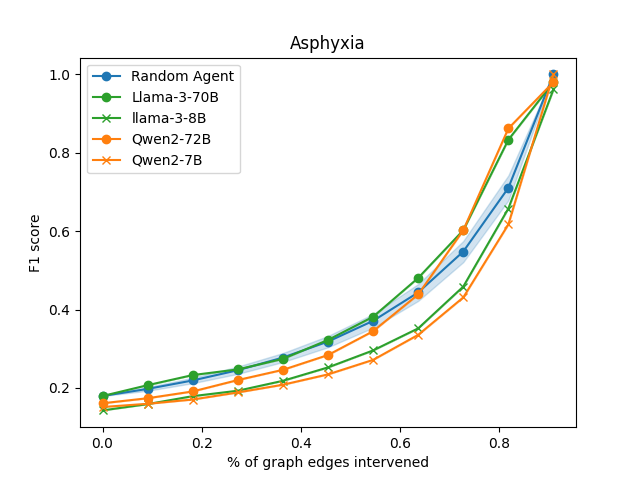}
    
    \includegraphics[width=0.32\linewidth]{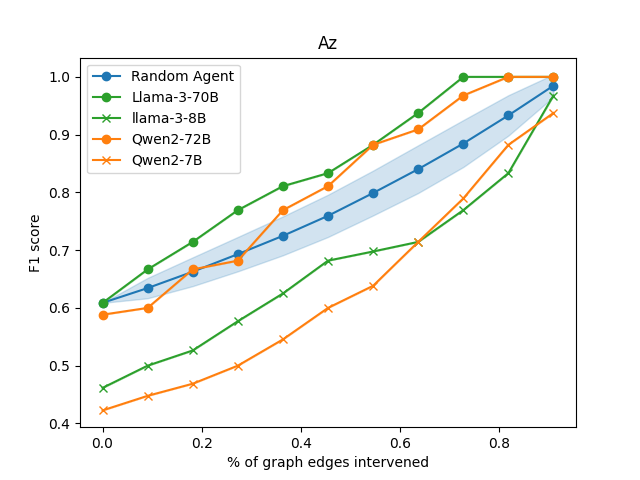}
    \includegraphics[width=0.32\linewidth]{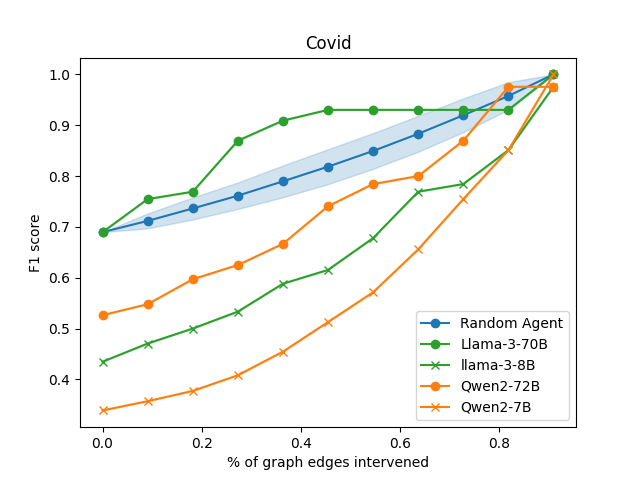}
    \includegraphics[width=0.32\linewidth]{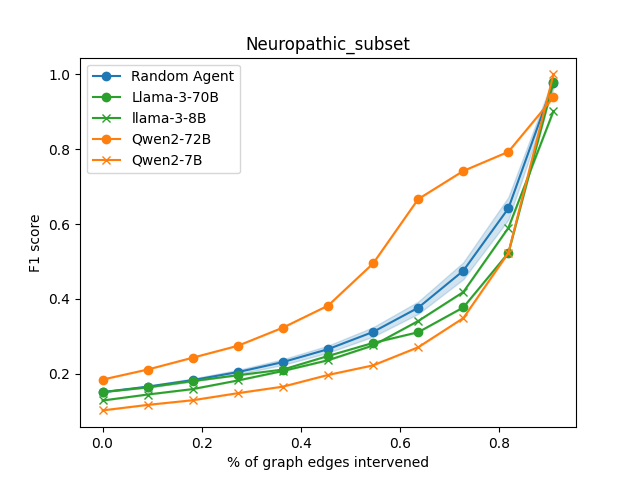}
    
    \includegraphics[width=0.32\linewidth]{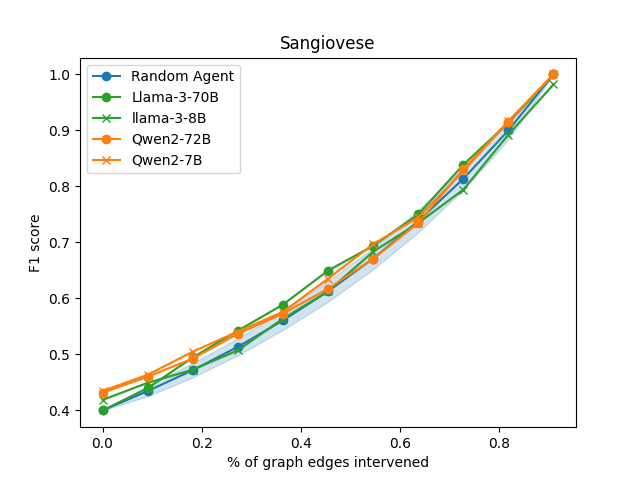}
    \caption{Static confidence selection over multiple models.}
    \label{fig:f1_static_confidence}
\end{figure}

\begin{figure}[ht]
    \centering
    \includegraphics[width=0.5\linewidth]{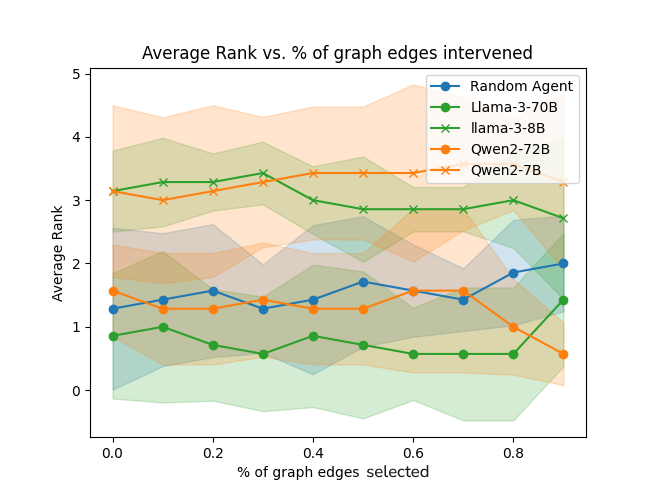}
    \caption{Static confidence based selection ranks for different models averaged across   graphs. \texttt{
Meta-Llama-3-70B-Instruct} is the only model to consistently outperform random guessing. Note: \textbf{lower is better}.}
    \label{fig:static_ranks}
\end{figure}

Figure \ref{fig:f1_static_confidence} reports the results of applying static confidence experiment selection using various models. Figure \ref{fig:static_ranks} reports the average rank for each model across benchmarked graphs.

\newpage

\section{Causal Graphs}
\label{sec:graphs}

Visualizations of causal graph benchmarks.

\begin{figure}[ht]
    \centering
    \includegraphics[width=0.7\linewidth]{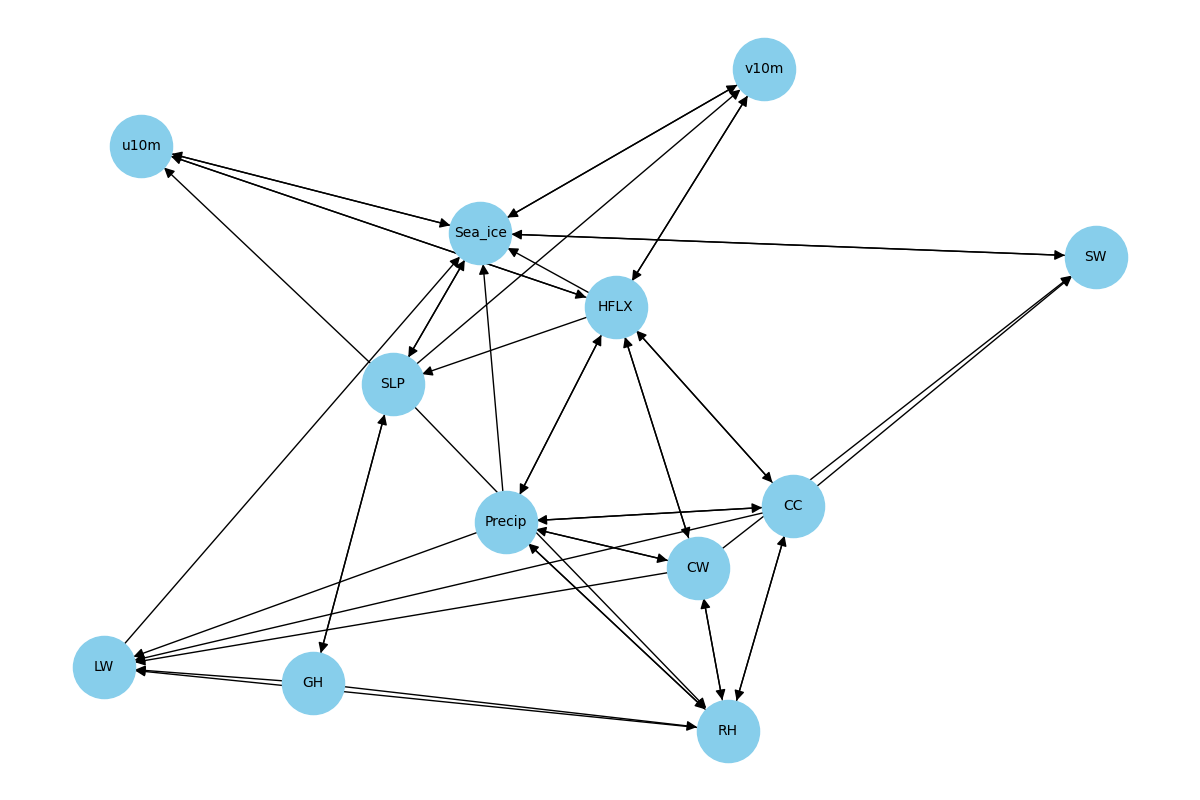}
    \caption{Arctic sea ice causal graph.}
    \label{graph:arctic}
\end{figure}

\begin{figure}
    \centering
    \includegraphics[width=0.7\linewidth]{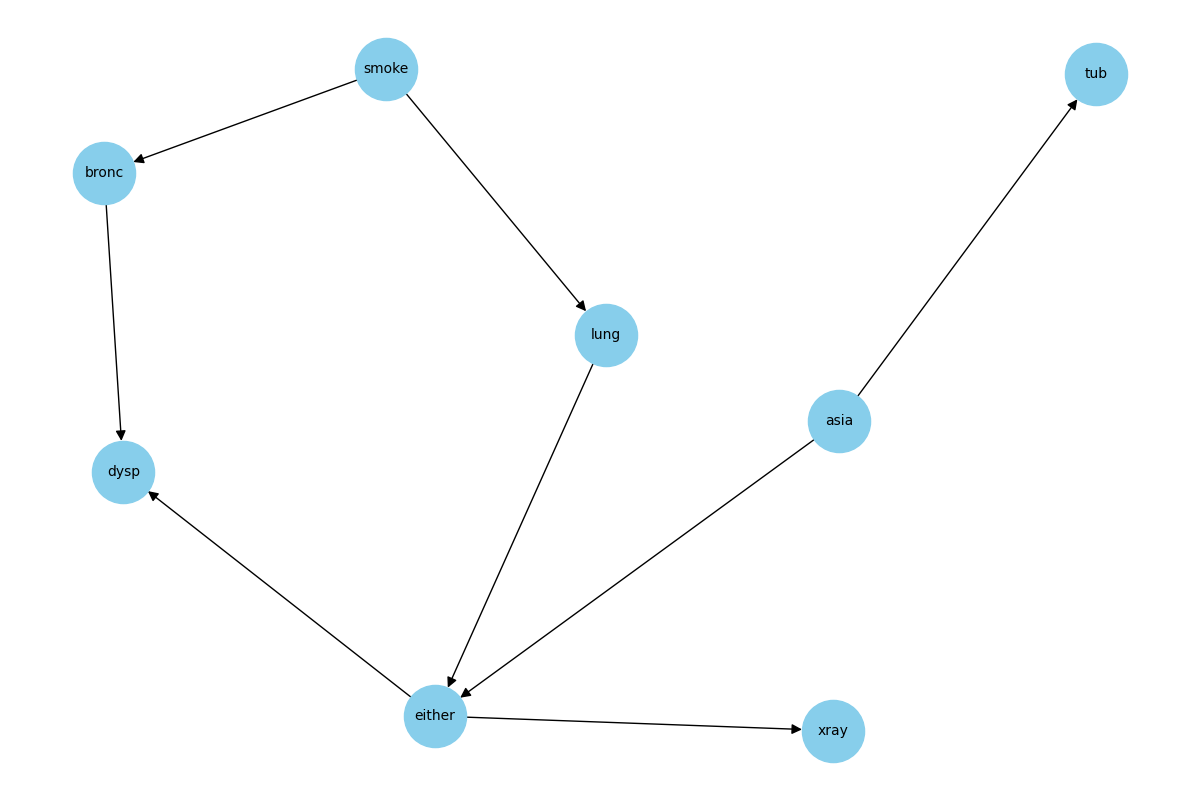}
    \caption{Asia causal graph.}
    \label{graph:asia}
\end{figure}

\begin{figure}
    \centering
    \includegraphics[width=0.7\linewidth]{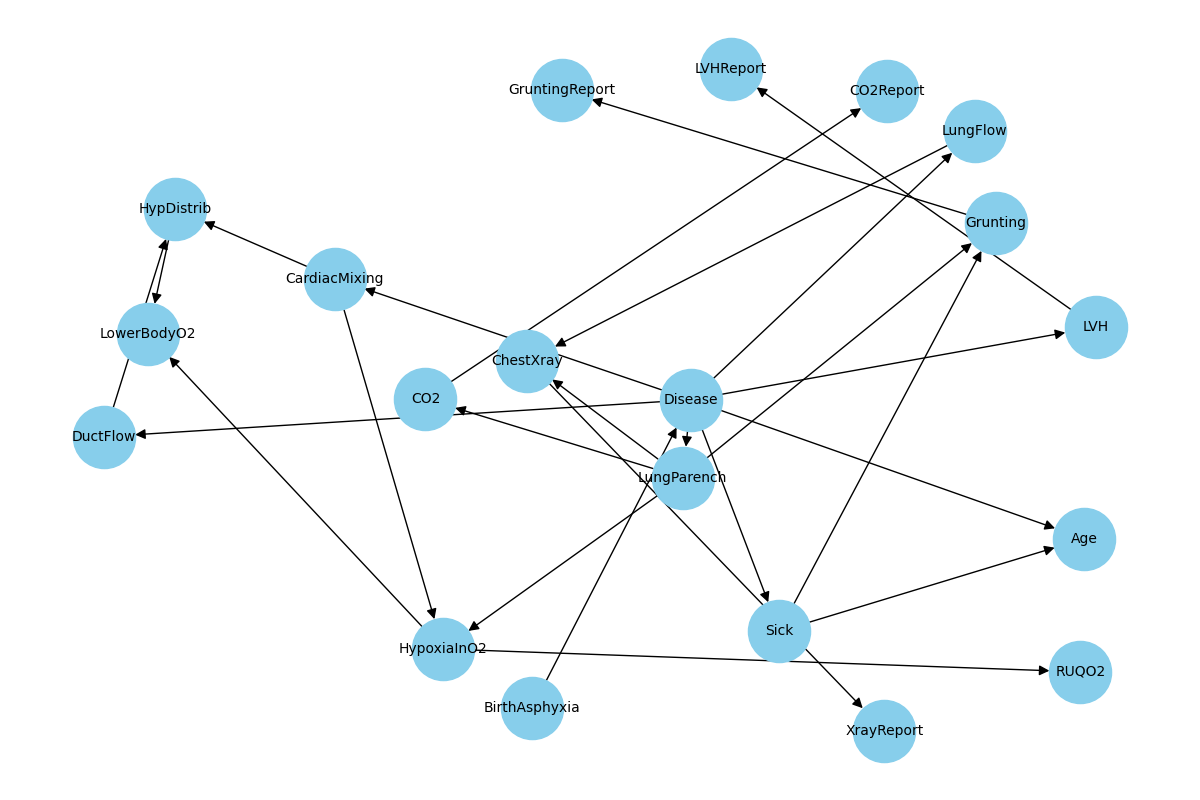}
    \caption{Asphyxia causal graph.}
    \label{graph:asphyxia}
\end{figure}

\begin{figure}
    \centering
    \includegraphics[width=0.7\linewidth]{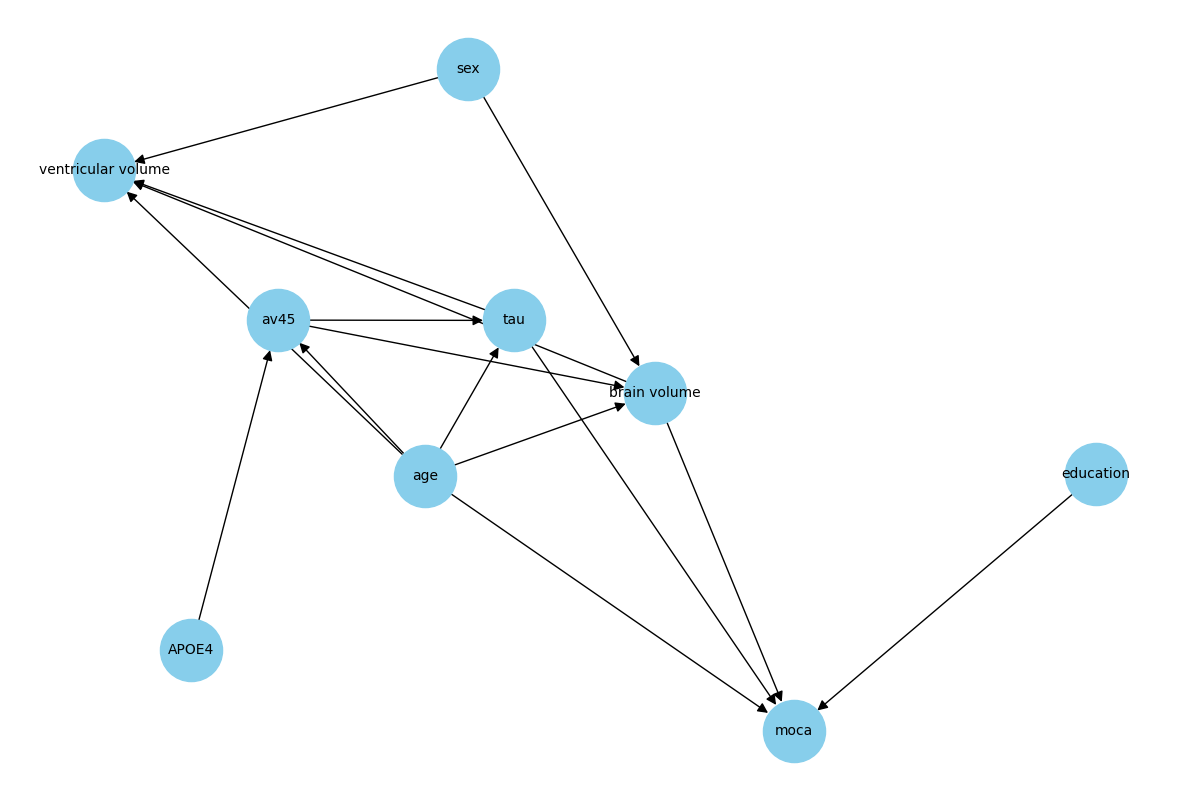}
    \caption{Alzheimers causal graph.}
    \label{graph:az}
\end{figure}

\begin{figure}
    \centering
    \includegraphics[width=0.7\linewidth]{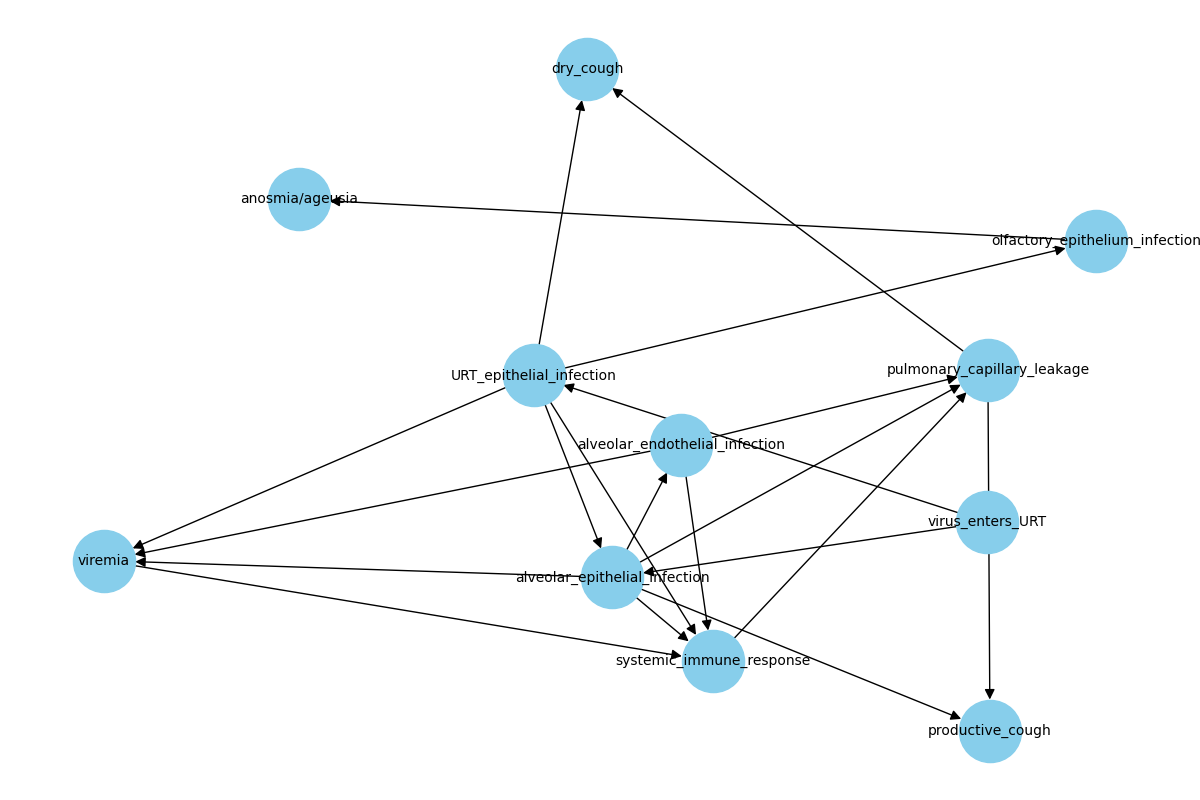}
    \caption{Covid causal graph.}
    \label{graph:covid}
\end{figure}

\begin{figure}
    \centering
    \includegraphics[width=0.7\linewidth]{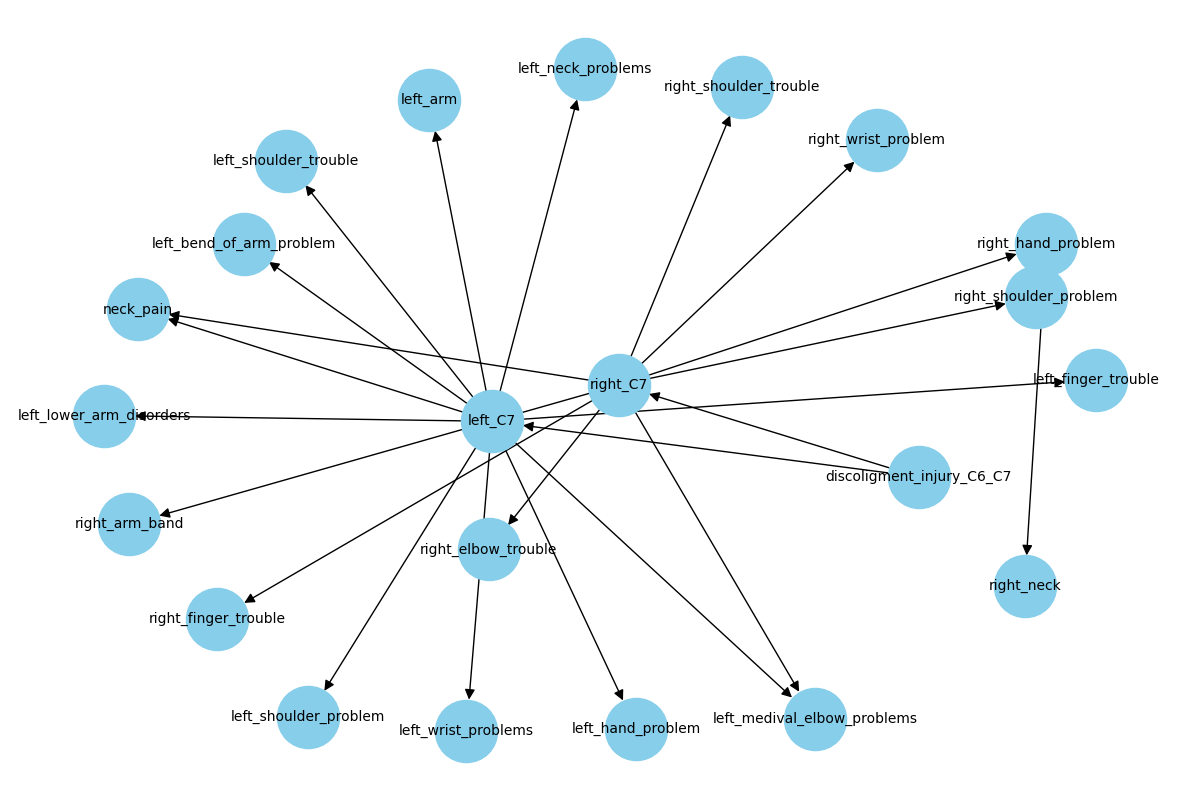}
    \caption{Neuropathic pain causal graph.}
    \label{graph:neuropathic}
\end{figure}

\begin{figure}
    \centering
    \includegraphics[width=0.7\linewidth]{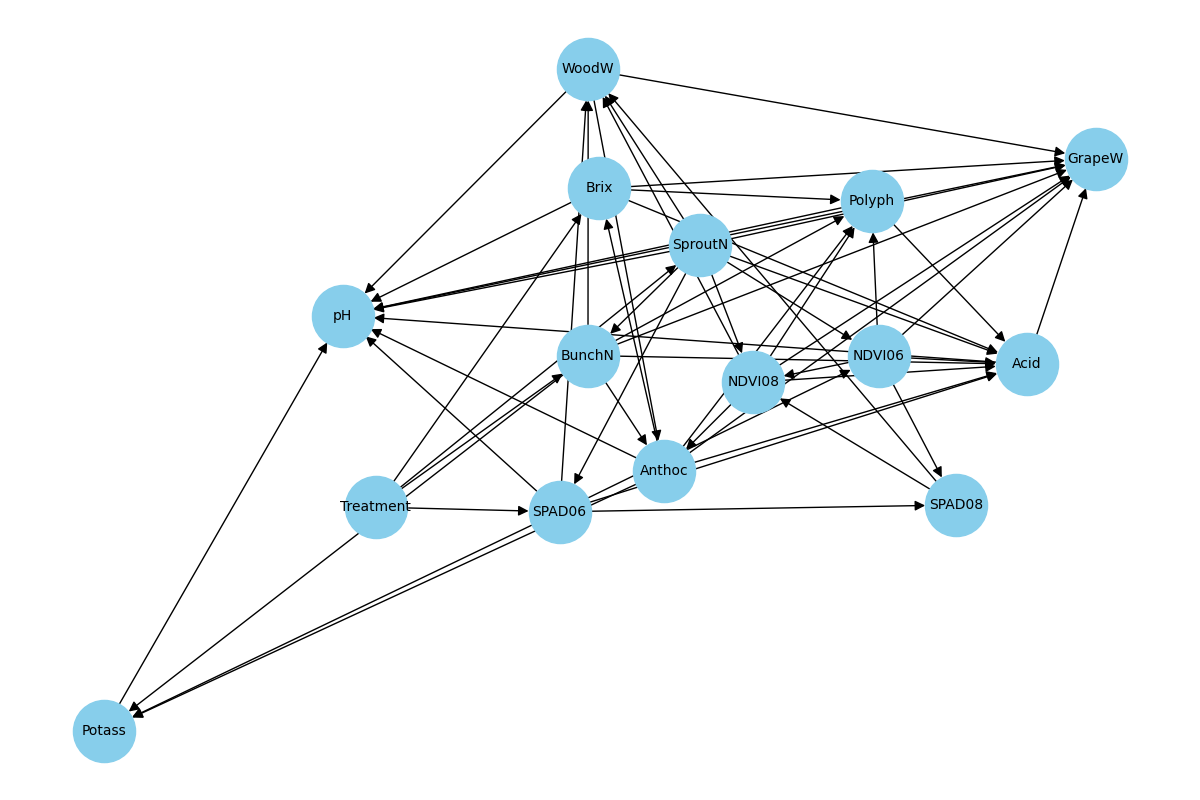}
    \caption{Sangiovese causal graph.}
    \label{graph:sangiovese}
\end{figure}

\begin{figure}
    \centering
    \includegraphics[width=0.7\linewidth]{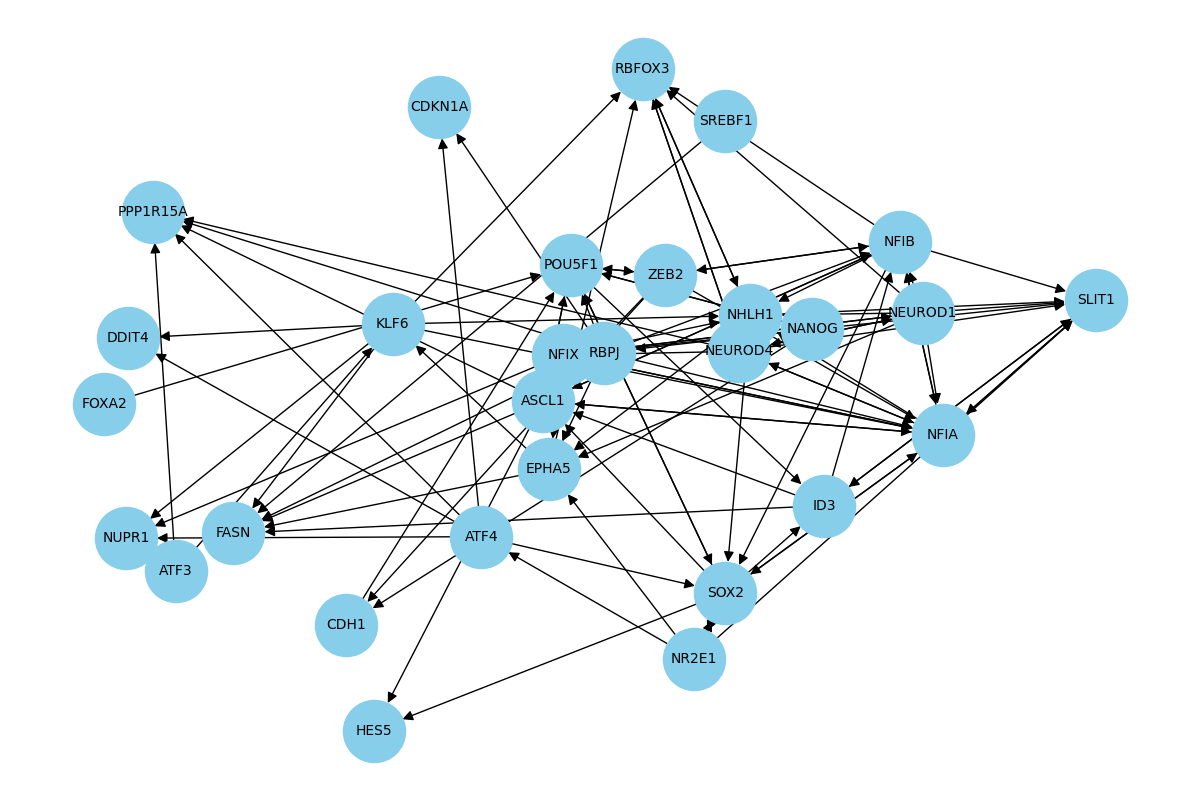}
    \caption{Brain causal graph.}
    \label{graph:brain}
\end{figure}

\newpage

\end{document}